\documentclass[lettersize,journal]{IEEEtran}
\usepackage{amsmath,amsfonts}
\usepackage{algorithmic}
\usepackage{array}
\usepackage[caption=false,font=normalsize,labelfont=sf,textfont=sf]{subfig}
\usepackage{textcomp}
\usepackage{stfloats}
\usepackage{url}
\usepackage{hyperref}
\usepackage{verbatim}
\usepackage{graphicx}
\usepackage{booktabs}
\usepackage{xcolor}
\usepackage{pifont}
\usepackage{subcaption}

\def\BibTeX{{\rm B\kern-.05em{\sc i\kern-.025em b}\kern-.08em
    T\kern-.1667em\lower.7ex\hbox{E}\kern-.125emX}}
\usepackage{balance}

\newcommand{\cmark}{\textcolor{green!60!black}{\ding{51}}}
\newcommand{\xmark}{\textcolor{red}{\ding{55}}}

\begin{document}
\title{OptiSight: Bridging Semantic Reasoning and Geometric Control for Embodied Navigation}

\author{Alperen Avan \quad Jordi Sanchez-Riera \\ 
Institut de Rob\`otica i Inform\`atica Industrial (CSIC-UPC)
\thanks{Manuscript created October, 2020; This work was developed by the IEEE Publication Technology Department. This work is distributed under the \LaTeX \ Project Public License (LPPL) ( http://www.latex-project.org/ ) version 1.3. A copy of the LPPL, version 1.3, is included in the base \LaTeX \ documentation of all distributions of \LaTeX \ released 2003/12/01 or later. The opinions expressed here are entirely that of the author. No warranty is expressed or implied. User assumes all risk.}}

\markboth{}
{How to Use the IEEEtran \LaTeX \ Templates}

\maketitle

\begin{abstract}
Autonomous indoor navigation requires both semantic understanding and precise geometric control. We propose OptiSight, a hybrid framework that combines Vision-Language Model reasoning with deterministic visual servoing through a finite-state Chain-of-Thought architecture. Grounded-SAM localizes open-vocabulary targets, while camera projection geometry converts visual observations into navigation commands without requiring dense mapping. The VLM is queried only at key decision points, reducing computational overhead while geometric control handles continuous navigation. Experiments in AI Habitat demonstrate reliable zero-shot navigation across diverse indoor scenarios, including obstacle avoidance and semantic ambiguity, while operating within an 8~GB VRAM budget. The source code is available at \url{https://github.com/avanalperen/OptiSight-Python-Multimodal-CoT-for-Visual-Reasoning}.

\end{abstract}


\section{Introduction}

Autonomous navigation in complex, unstructured indoor environments remains a fundamental challenge in robotics and embodied artificial intelligence. Beyond reaching a target location while avoiding obstacles, intelligent agents must understand the semantic structure of their surroundings to execute high-level tasks such as locating a doorway, identifying a navigable passage, or searching for a previously unseen object. Achieving this capability requires combining geometric navigation with semantic scene understanding, enabling robots to reason about both the physical layout and the meaning of the environment.

Traditional navigation systems are predominantly built upon Simultaneous Localization and Mapping (SLAM) and classical geometric path-planning algorithms. These methods have demonstrated remarkable success in localization, map construction, and collision-free navigation while remaining computationally efficient enough for real-time operation. However, because they rely primarily on geometric representations, their ability to interpret semantic information is inherently limited. Consequently, they struggle with tasks involving open-vocabulary object recognition, contextual scene understanding, and natural-language instructions, making them unsuitable for many embodied AI applications.

Recent advances in Vision-Language Models (VLMs) have significantly expanded the semantic capabilities of robotic systems. Trained on large-scale multimodal datasets, these models provide powerful zero-shot perception and semantic reasoning capabilities, enabling robots to recognize previously unseen objects, understand contextual relationships, and interpret natural-language instructions without task-specific retraining. These advances have brought embodied agents closer to human-like scene understanding and represent an important step toward more intelligent autonomous systems.

\begin{figure}[t]
    \centering
    \includegraphics[width=0.49\textwidth]{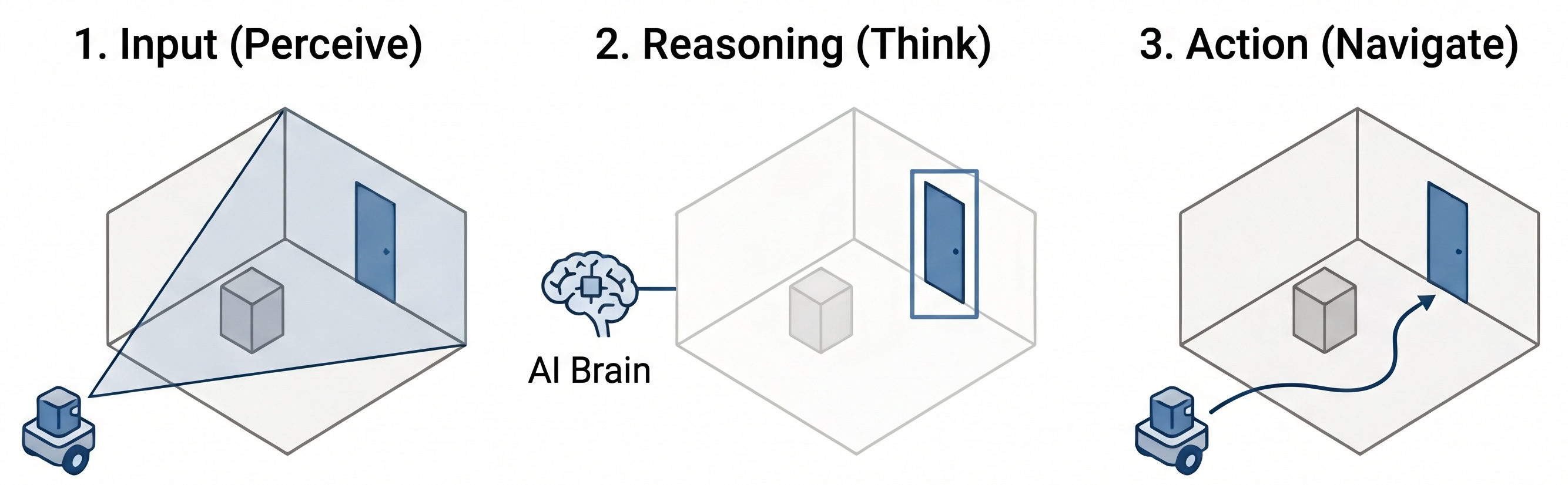}
    \caption{Conceptual overview of the OptiSight framework. The system perceives the environment, isolates the target via VLM reasoning, and executes a smooth geometric path.}
    \label{fig:conceptual}
\end{figure}

Despite these capabilities, VLMs are not designed to directly control robotic platforms. While they excel at interpreting visual scenes and reasoning about semantic concepts, they do not naturally translate high-level goals into executable navigation strategies or precise motion commands. Furthermore, state-of-the-art VLMs impose substantial computational and memory requirements, resulting in inference latencies that are often incompatible with real-time control on resource-constrained platforms. Even when a VLM correctly infers a semantic concept—for example, identifying the doorway that leads outside a room—it lacks the geometric precision required to generate collision-free trajectories and continuous motion commands.

Consequently, neither purely geometric navigation nor purely semantic reasoning alone provides a complete solution for autonomous embodied navigation. An effective navigation framework must combine the semantic understanding and reasoning capabilities of VLMs with the efficiency, precision, and robustness of classical geometric navigation. Moreover, it should decompose abstract user instructions, such as \emph{"get out of the room"}, into a sequence of intermediate semantic objectives that can be safely executed by a low-level navigation controller while remaining computationally lightweight enough for deployment on edge devices and wearable robotic platforms.

To address these challenges, we propose \textbf{OptiSight}, a hybrid autonomous navigation framework that bridges semantic understanding and geometric navigation through three complementary stages: \emph{Perceive}, \emph{Think}, and \emph{Navigate} (Fig.~\ref{fig:conceptual}). Given a high-level instruction, OptiSight first perceives the surrounding environment, then reasons about the semantic actions required to accomplish the task, and finally translates these decisions into precise geometric control commands for robust real-time navigation.

Operating within the AI Habitat simulation environment, OptiSight first perceives the surrounding scene, then employs a Vision-Language Model to identify and reason about semantic targets, and finally converts these high-level observations into precise geometric navigation commands through mathematically grounded visual servoing. By combining semantic reasoning with deterministic geometric control, OptiSight enables robust and efficient navigation while remaining suitable for real-time deployment on computationally constrained platforms.

A key component of OptiSight is a semantic-to-geometric translation module that directly converts semantic observations into navigation actions. Existing embodied navigation approaches typically address this problem either through end-to-end reinforcement learning or by constructing dense semantic scene representations. Reinforcement learning methods often suffer from poor sim-to-real transfer and require extensive task-specific training, whereas semantic SLAM and neural scene representation techniques incur substantial computational and memory costs that limit their deployment on lightweight hardware. Instead, OptiSight leverages Grounded-SAM~\cite{Ren2024GroundedSAM} to identify open-vocabulary targets and extracts their semantic boundaries in the image plane. Using differentiable camera projection geometry, these observations are transformed directly into perspective-correct navigation angles $(\theta,\alpha)$, enabling precise closed-loop visual servoing without requiring dense mapping or neural scene reconstruction.

A second challenge lies in efficiently integrating VLM reasoning into the navigation loop. Many recent embodied AI systems continuously query the VLM throughout navigation, introducing substantial inference latency due to repeated multimodal reasoning. This frequent interaction not only limits real-time responsiveness but also increases computational and memory requirements, making deployment on edge devices impractical. OptiSight addresses this limitation through a state-driven Chain-of-Thought architecture based on a finite-state machine comprising \textit{Search}, \textit{Find}, \textit{Scan}, \textit{Navigate}, and \textit{Recover} states. Rather than invoking the VLM continuously, semantic reasoning is performed only at critical state transitions, while deterministic visual servoing governs continuous low-level motion. This separation between high-level cognition and low-level control substantially reduces inference overhead while preserving explainable reasoning and real-time navigation performance.

Finally, we validate OptiSight through extensive experiments in AI Habitat~\cite{ai_habitat} using the Habitat-Sim simulator~\cite{habitat_sim}. We demonstrate that the proposed framework achieves reliable semantic navigation in complex indoor environments while maintaining the computational efficiency required for deployment on resource-constrained robotic platforms.

\section{Related work}

\subsection{Vision-Language Models for Embodied Navigation}

The rapid progress of Large Language Models (LLMs) and Vision-Language Models (VLMs) has significantly advanced embodied artificial intelligence by enabling robots to reason about complex environments using natural language instructions. Rather than relying solely on geometric representations, these models provide semantic understanding, open-vocabulary perception, and contextual reasoning, allowing agents to execute tasks beyond traditional navigation objectives.

Recent foundation models have demonstrated increasingly general embodied capabilities. HabitatGS~\cite{xia2026habitatgs} extends AI Habitat with photorealistic Gaussian Splatting scenes populated by dynamic humans, providing a more realistic benchmark for embodied reasoning. Embodied foundation models such as EmbodiedFM~\cite{zhang2026embodied} are trained on millions of navigation demonstrations spanning multiple robotic domains—including household robots, drones, and autonomous vehicles—and demonstrate strong zero-shot generalization across diverse navigation tasks.

Several works have also investigated reasoning-driven navigation. OctoNav~\cite{gao2025octonav} introduces a think-before-action paradigm in which navigation policies are trained from instruction-trajectory pairs using reinforcement learning. NavR1~\cite{liu2025navr1} further incorporates Group Relative Policy Optimization (GRPO) together with Chain-of-Thought supervision to jointly learn dialogue, planning, reasoning, and navigation. These methods demonstrate that explicit reasoning improves embodied decision making, although they generally require extensive training datasets and computationally expensive foundation models.

Unlike these approaches, OptiSight does not require end-to-end policy learning or large-scale navigation datasets. Instead, it leverages pretrained VLMs only for semantic perception and high-level reasoning, while delegating continuous motion generation to deterministic geometric control.

\subsection{Semantic and Object-Goal Navigation}

Object-goal navigation has emerged as a fundamental benchmark for embodied AI, requiring agents to locate objects specified by semantic categories rather than predefined coordinates. OVON~\cite{yokoyama2024ovon} significantly expanded this setting by introducing over 15,000 object instances and demonstrating open-vocabulary object navigation using semantic representations.

Subsequent work has focused on improving navigation efficiency through memory-augmented reasoning. EfficientNav~\cite{yang2026efficientnav} combines zero-shot LLM planning with semantic memory retrieval and memory clustering to reduce repeated reasoning and inference latency. Hierarchical reasoning frameworks such as HiRobot~\cite{shi-icml-hirobot} decompose complex user instructions into hierarchical subgoals while incorporating user feedback to refine navigation plans.

Although these approaches improve semantic navigation, they typically rely on repeated LLM or VLM inference throughout execution, increasing computational overhead. In contrast, OptiSight invokes semantic reasoning only at critical decision points while maintaining continuous navigation through lightweight visual servoing, substantially reducing latency.

\subsection{Semantic Grounding and Chain-of-Thought Reasoning}

A growing body of work has explored explicit reasoning mechanisms for embodied agents. Chain-of-Thought (CoT) prompting was introduced by Wei \emph{et al.}~\cite{wei_cot}, demonstrating that explicitly generating intermediate reasoning steps substantially improves the ability of large language models to solve complex multi-step problems. Building upon this foundation, recent research has extended CoT reasoning to robotics by integrating semantic reasoning with perception and control. For example, Shen \emph{et al.}~\cite{shen_enhancing} proposed a multimodal CoT framework for collaborative multi-robot semantic navigation, while Wang \emph{et al.}~\cite{wang_deductive} incorporated deductive CoT reasoning into socially-aware navigation world models. These works highlight the growing importance of structured reasoning for embodied intelligence.

More recent methods integrate CoT directly into embodied navigation pipelines. NavCoT~\cite{lin2024navcot}, CoTVLA~\cite{zhao2025cotvla}, UniNaVid~\cite{zhang2024uninavid}, and FantasyVLN~\cite{zuo2026fantasyvln} incorporate explicit reasoning into navigation policies to improve long-horizon planning. Dynamic 3D-VLP~\cite{wang2026d3dvlp} further proposes a unified 3D Chain-of-Thought framework that jointly performs planning, grounding, navigation, and question answering within a single 3D vision-language model. In parallel, SceneGraph-VLM-based approaches~\cite{makarov2026scenegraphvlm,SceneGraph_luca,loo2025open,rajvanshi2024saynav,hu2025imaginative,huang2025msgnav,zawalski2024robotic} exploit structured scene graphs to improve semantic grounding and spatial reasoning.

Beyond navigation, similar reasoning paradigms have also been explored for robotic manipulation. ALRM~\cite{santos2026alrm} combines ReAct-style reasoning with executable code generation and tool-based planning to perform long-horizon manipulation tasks.

While these methods demonstrate the benefits of explicit reasoning, many continuously query large VLMs throughout task execution, resulting in substantial computational cost and inference latency. In contrast, OptiSight adopts a state-driven Chain-of-Thought architecture in which semantic reasoning is invoked only at key state transitions within a finite-state machine, while continuous low-level motion is handled by deterministic visual servoing. This separation between high-level reasoning and geometric control enables explainable navigation while maintaining real-time performance on resource-constrained robotic platforms.

\section{Methodology}

\begin{figure*}[t]
    \centering
    \includegraphics[width=0.95\textwidth]{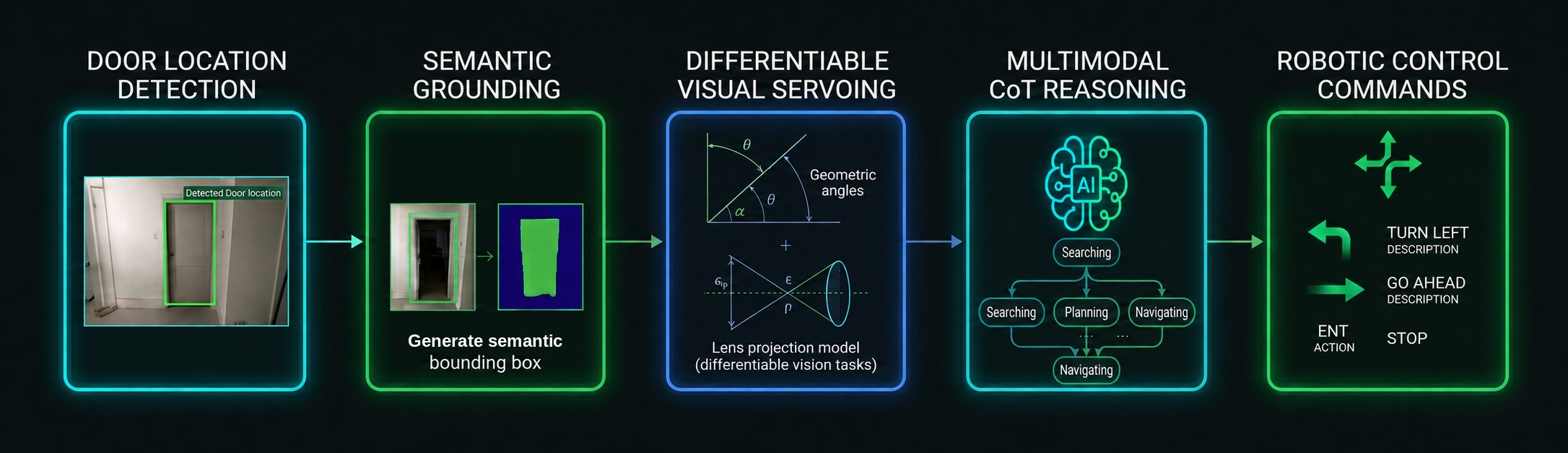}
    \caption{Conceptual overview of the OptiSight framework. The system perceives the environment, isolates the target via VLM reasoning, and executes a smooth geometric path.}
    \label{fig:architecture}
\end{figure*}

OptiSight is a closed-loop autonomous navigation framework that integrates Vision-Language Model (VLM) reasoning, open-vocabulary object grounding, geometric projection, and deterministic motion control. In its current implementation, the system is designed to execute the high-level navigation instruction \emph{"Get out of the room"}. To accomplish this task, the framework decomposes the instruction into a sequence of semantic reasoning and geometric navigation stages, as illustrated in Figure~\ref{fig:architecture}.

The execution flow is governed by a finite-state machine (FSM) comprising five states: \textit{Search}, \textit{Find}, \textit{Scan}, \textit{Navigate}, and \textit{Recover}. Each state encapsulates a specific subtask and defines both the required perception module and the transition conditions to subsequent states. Rather than continuously querying the VLM throughout execution, OptiSight invokes it only at semantic decision points. Each state is associated with a dedicated prompt template that provides task-specific context, allowing the FSM itself to maintain the execution context and eliminating the need for long conversational histories. During the \textit{Search} state, the VLM analyzes the current scene to determine whether the semantic target is visible and decides the next navigation action. Once the target has been identified, the framework transitions to deterministic geometric processing, and the VLM is not queried again unless the system enters the \textit{Recover} state.

The nominal execution pipeline follows the sequence \textit{Search} $\rightarrow$ \textit{Find} $\rightarrow$ \textit{Scan} $\rightarrow$ \textit{Navigate}. In the \textit{Find} stage, Grounded-SAM~\cite{Ren2024GroundedSAM} localizes the target object using open-vocabulary segmentation. The segmented target is then projected into the 3D camera reference frame to estimate its spatial location. Subsequently, during the \textit{Scan} stage, the camera is automatically tilted downward by $40^{\circ}$ to analyze the traversable floor area and detect potential obstacles. Based on the resulting geometric representation, the planner computes either a direct trajectory toward the target or a collision-free bypass trajectory around detected obstacles.

During the \textit{Navigate} state, the camera maintains its downward orientation to continuously monitor the planned path while proportional steering commands drive the robot along a sequence of 3D waypoints. Waypoints are generated from the planned trajectory and are sequentially discarded once the robot approaches within $0.30$\,m of the current target waypoint, ensuring smooth trajectory tracking and stable motion execution. If the semantic target is lost or the planned path becomes invalid, the FSM transitions to the \textit{Recover} state, where the VLM is invoked again to re-establish the navigation objective before resuming the execution pipeline. 

The following subsections describe the operations performed in each state of the finite-state machine and the actual implementation of the proposed method.

\subsection{Finite State Machine (FSM)}

\begin{figure}[t]
    \centering
    \includegraphics[width=0.49\textwidth]{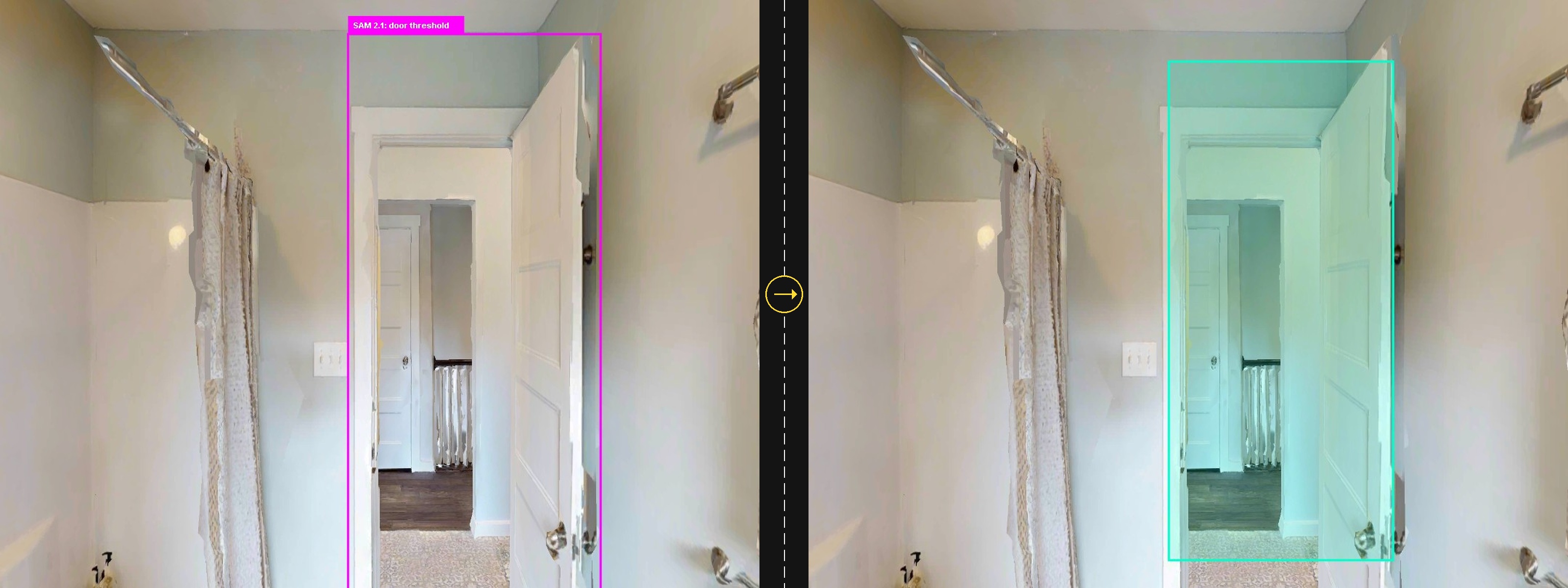}
    \caption{(Step Find) Semantic object localization and segmentation for a given object target.}
    \label{fig:search}
\end{figure}

\textbf{Search.} The \textit{Search} state is responsible for identifying the semantic target specified by the navigation task, see Figure~\ref{fig:search}. The current RGB image is provided to the Vision-Language Model (VLM), which determines whether the target (e.g., a doorway) is visible and recommends the next navigation action. If the target is not detected, the robot executes an autonomous scanning procedure consisting of nine successive rotations of $10^{\circ}$, separated by $0.3$\,s intervals, allowing the camera to inspect the surrounding environment until the target enters the field of view.

\textbf{Find.} Once the target has been detected, the framework transitions to the \textit{Find} state, where Grounded-SAM~\cite{Ren2024GroundedSAM} performs open-vocabulary localization to obtain an accurate segmentation of the target. To improve localization stability, the segmentation is refined through up to three consecutive inference passes, producing a consistent low-noise bounding region. The resulting image coordinates are then projected into the 3D camera reference frame using the geometric projection model, yielding the spatial location of the navigation target. Upon successful projection, the system proceeds to the \textit{Scan} state.

\begin{figure}[t]
    \centering
    \includegraphics[width=0.49\textwidth]{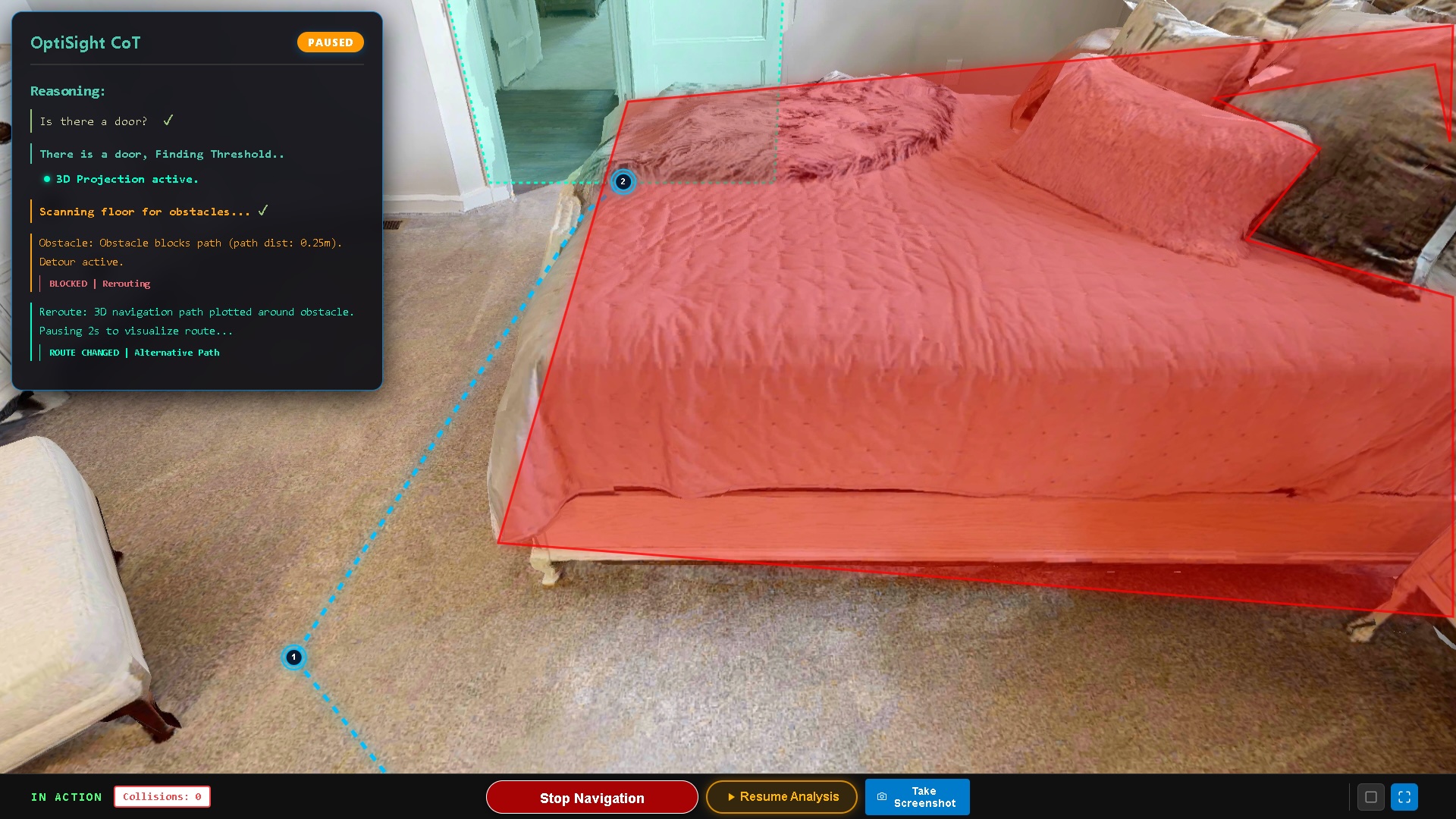}
    \caption{(Step Scan) Obstacle detection given waypoint path to achieve final goal.}
    \label{fig:scan}
\end{figure}

\textbf{Scan.} During the \textit{Scan} state, the camera is automatically tilted downward by $40^{\circ}$ to maximize visibility of the traversable floor area. Grounded-SAM is then employed to segment potential obstacles, whose image coordinates are projected into 3D space to construct a local obstacle representation. If an obstacle lies within a predefined safety margin of $0.50$\,m, the planner generates a collision-free bypass trajectory around the obstacle while maintaining continuous visual monitoring through the downward-facing camera, see Figure~\ref{fig:scan}. Otherwise, a direct trajectory toward the target is computed and the camera is restored to its nominal horizontal orientation. This procedure enables deterministic obstacle avoidance without requiring repeated VLM inference during navigation.

\begin{figure}[t]
    \centering
    \includegraphics[width=0.49\textwidth]{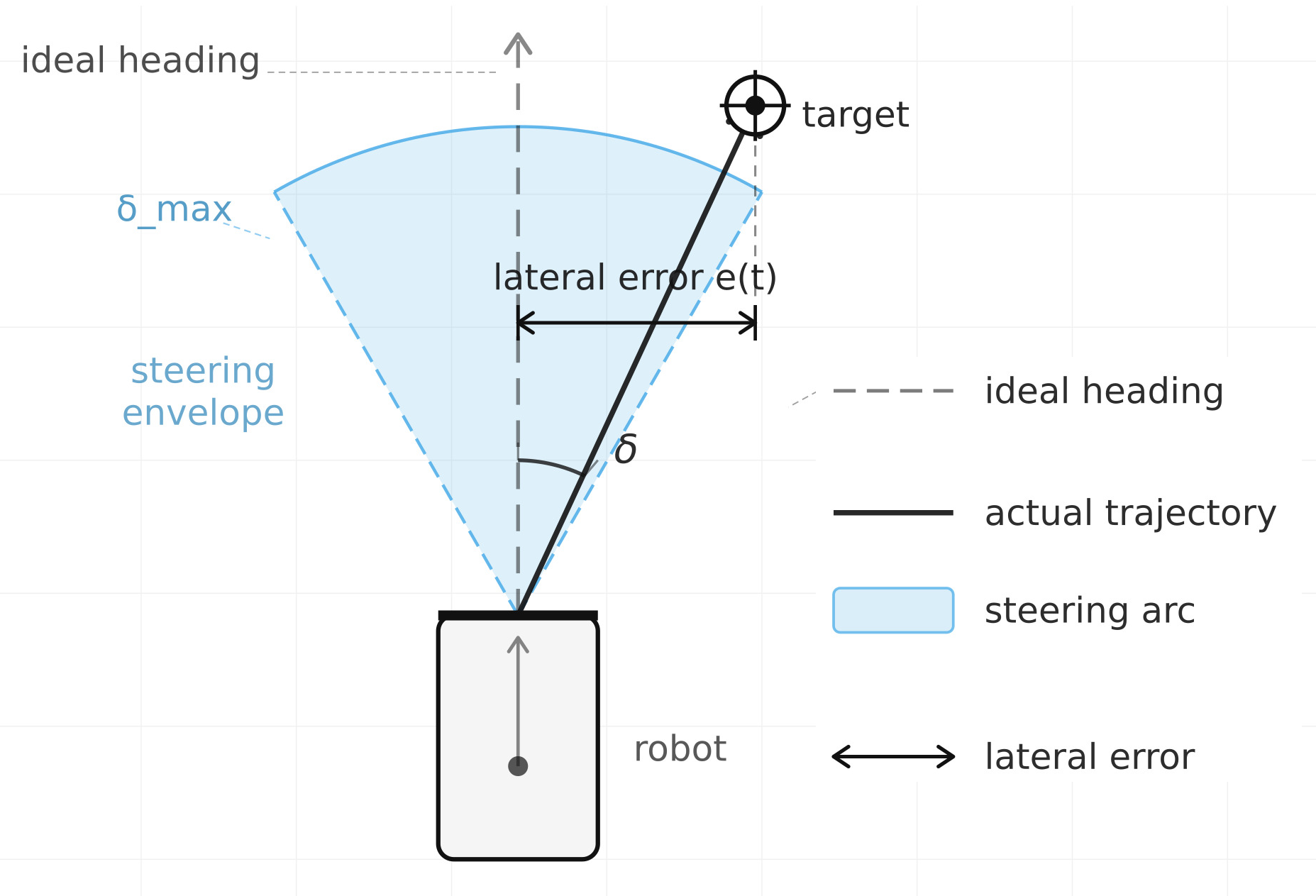}
    \caption{(Step Navigate) Steering calculation towards the target.}
    \label{fig:navigate}
\end{figure}

\textbf{Navigate.} After a valid trajectory has been generated, the FSM transitions to the \textit{Navigate} state. The robot follows the planned trajectory through a sequence of 3D waypoints using proportional visual servoing. At each control cycle, the controller computes the lateral deviation of the active waypoint with respect to the camera center and generates steering commands scaled according to the waypoint's projected size, ensuring stable motion under the simulator's kinematic constraints, see Figure~\ref{fig:navigate}. A waypoint is considered reached once the Euclidean distance between the robot and the waypoint falls below $0.30$\,m, after which the controller automatically switches to the next waypoint. To account for dynamic changes in the environment, the navigation process is periodically interrupted every $15$ control steps, returning the FSM to the \textit{Scan} state to reassess floor traversability and update the planned trajectory if new obstacles have appeared.

\subsection{Dual-Isolated System Architecture}

OptiSight adopts a dual-isolated execution architecture that separates semantic reasoning from geometric navigation into two independent runtime environments. The two environments communicate through a lightweight socket-based interface, allowing each component to operate within its own software stack while exchanging only the information required for navigation.

The first environment executes Habitat-Sim together with the geometric navigation pipeline, including the finite-state machine, camera control, geometric projection, path planning, waypoint generation, and motion control. The second environment hosts the Vision-Language Model (VLM) and the Grounded-SAM inference pipeline responsible for semantic reasoning and open-vocabulary object localization.

Communication between both environments is asynchronous. Whenever semantic reasoning is required, the navigation module sends the current observation and task prompt to the VLM server, which performs inference independently and returns the resulting semantic prediction. Since the VLM is queried only at the semantic decision points defined by the finite-state machine, geometric navigation continues uninterrupted while reasoning is executed in the background. Once the response is received, the finite-state machine resumes execution using the inferred semantic information.

This architecture isolates the dependencies of the robotics and VLM frameworks, enabling the use of modern multimodal models together with legacy robotics software without compatibility issues. In addition, separating semantic inference from geometric control improves resource utilization and allows the complete system to operate within an 8,GB VRAM budget suitable for deployment on resource-constrained edge platforms such as NVIDIA Jetson~\cite{nvidia_jetson}.

\section{Experiments}

We evaluate OptiSight in the AI Habitat simulation environment across $24$ diverse indoor navigation scenarios. In each scenario, the agent is tasked with executing the high-level instruction \emph{"Get out of the room."} and is evaluated over $10$ independent runs. The first $12$ scenarios use Qwen3.5-2B~\cite{qwen35} for VLM reasoning, while the remaining $12$ scenarios use Moondream2-2B~\cite{vik_2024_moondream2}. Grounded-SAM is used throughout all experiments for open-vocabulary target localization.

The selected scenarios are designed to evaluate the robustness of OptiSight under six representative navigation challenges: (i) single-obstacle avoidance, (ii) dual-obstacle reasoning, (iii) semantic disambiguation between structurally similar objects, (iv) partial observability requiring active visual search, (v) extreme viewpoints producing geometric distortions, and (vi) perceptual ambiguities caused by reflective surfaces. The first three scenario types primarily evaluate navigation and semantic grounding, whereas the latter three focus on perception and geometric ambiguities that can challenge VLM-based reasoning. This set of scenarios therefore provides a diverse evaluation of both the semantic and geometric components of the proposed framework.

All experiments are executed using a fully automated evaluation pipeline. During each run, the system records quantitative navigation metrics together with structured step-by-step execution traces, including state transitions, VLM requests, parsing events, collisions, and recovery actions. For each scenario, the reported results correspond to the mean values computed over the $10$ independent runs. The results are summarized in Tables~\ref{tab:final_comparison_1} and~\ref{tab:final_comparison_2}.

We evaluate OptiSight using the following metrics:

\begin{itemize}
\item \textbf{Mission Success:} Binary indicator of whether the navigation objective is successfully completed. A scenario is considered successful if the mission is completed in at least half of the $10$ independent runs.
\item \textbf{Success Rate:} Percentage of successful runs across the $10$ trials.
\item \textbf{Execution Time:} Time required to complete the navigation task, measured in seconds.
\item \textbf{Total Distance:} Cumulative distance traveled by the agent, measured in meters.
\item \textbf{Total Steps:} Total number of recorded state transitions, actions, and movement steps during execution.
\item \textbf{Collisions:} Number of physical collisions with environmental obstacles.
\item \textbf{Recoveries:} Number of recovery or obstacle-avoidance procedures triggered during navigation.
\item \textbf{VLM Requests:} Number of VLM inference calls performed during a navigation episode.
\item \textbf{Minimum Obstacle Distance:} Minimum Euclidean distance, in meters, between the agent and any detected 3D obstacle during navigation.
\end{itemize}

Together, these metrics characterize both task-level performance and the internal behavior of the system. Success rate, execution time, and traveled distance measure navigation effectiveness, while collisions and minimum obstacle clearance quantify safety. VLM requests, parsing failures, and recovery actions provide additional measures of computational and operational robustness.

\subsection{Results}

\begin{figure*}[ht]
  \centering
  
  \subfloat[Exp 01]{\includegraphics[width=0.25\textwidth]{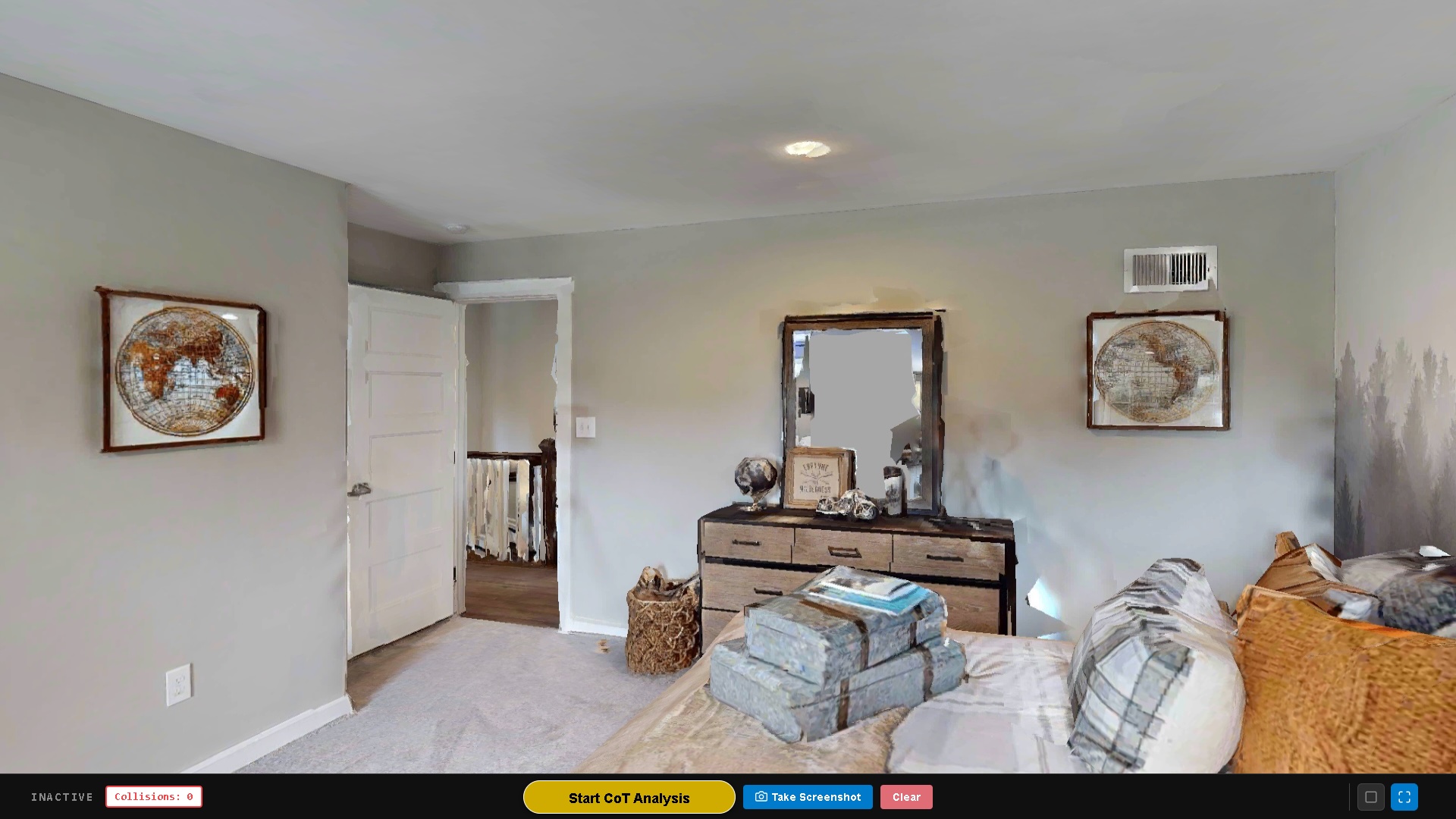}} 
  \subfloat[Exp 02]{\includegraphics[width=0.25\textwidth]{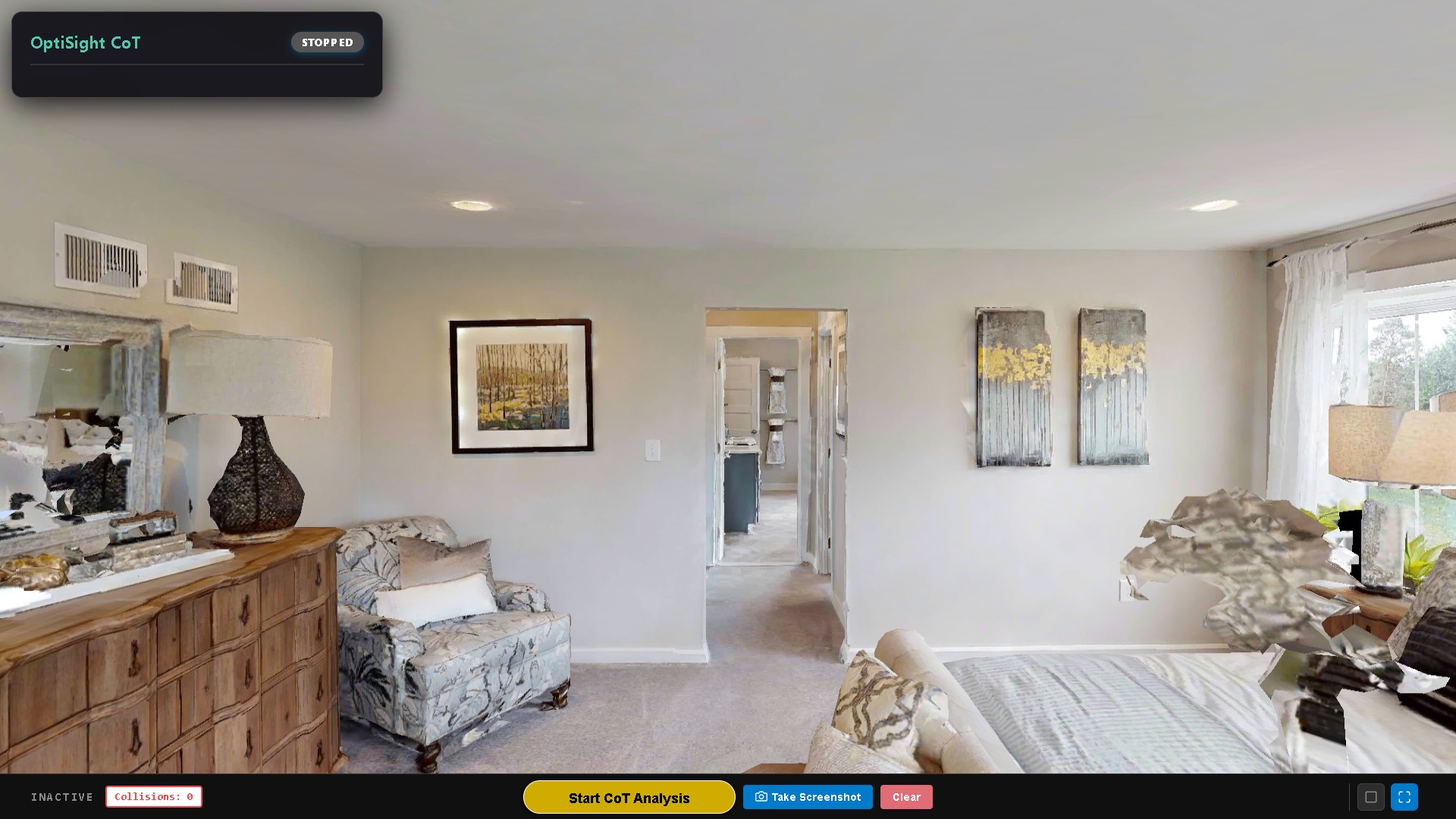}} 
  \subfloat[Exp 03]{\includegraphics[width=0.25\textwidth]{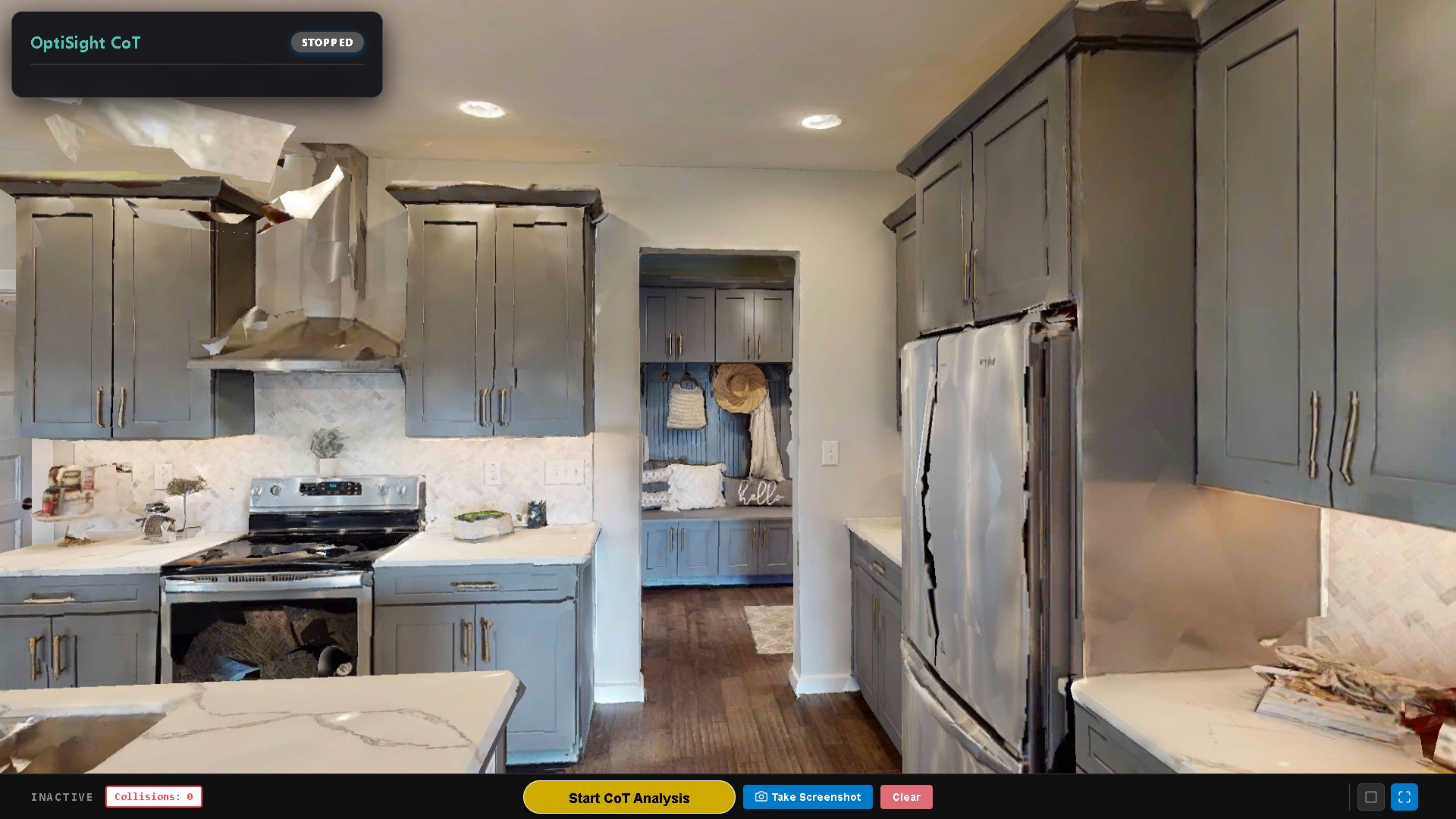}} 
  \subfloat[Exp 04]{\includegraphics[width=0.25\textwidth]{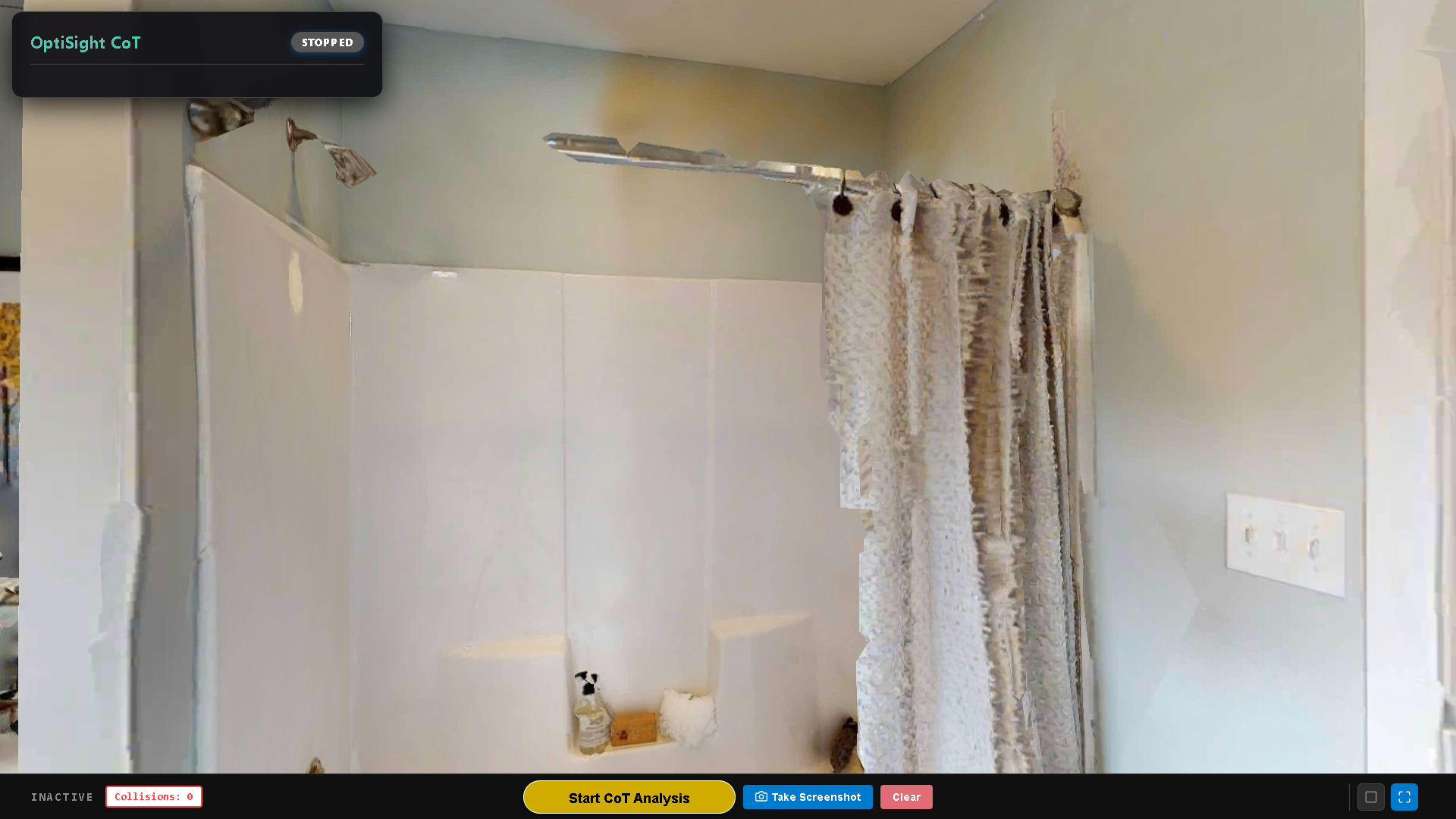}} 

  ~\\

  \subfloat[Exp 05]{\includegraphics[width=0.25\textwidth]{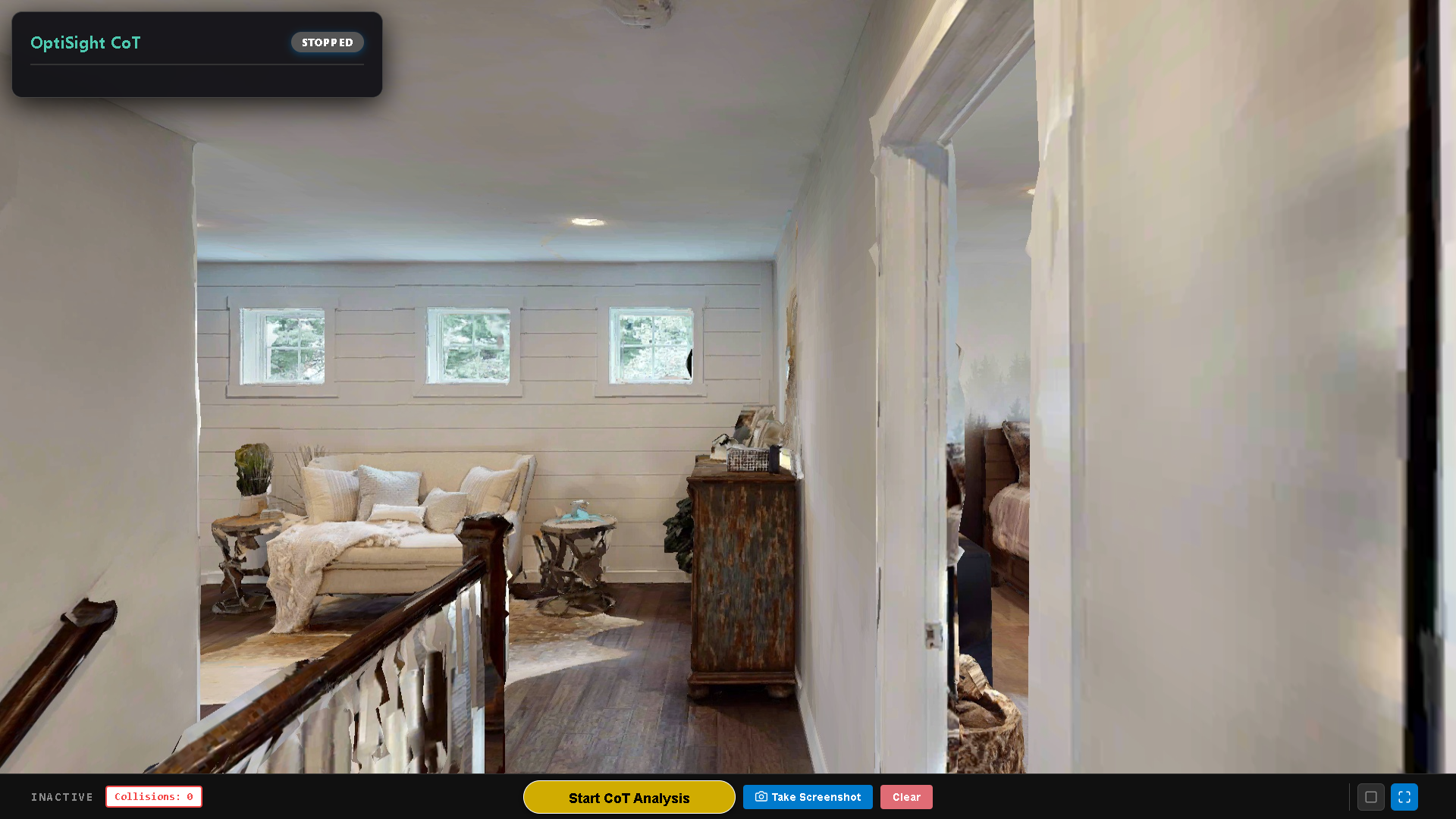}} 
  \subfloat[Exp 06]{\includegraphics[width=0.25\textwidth]{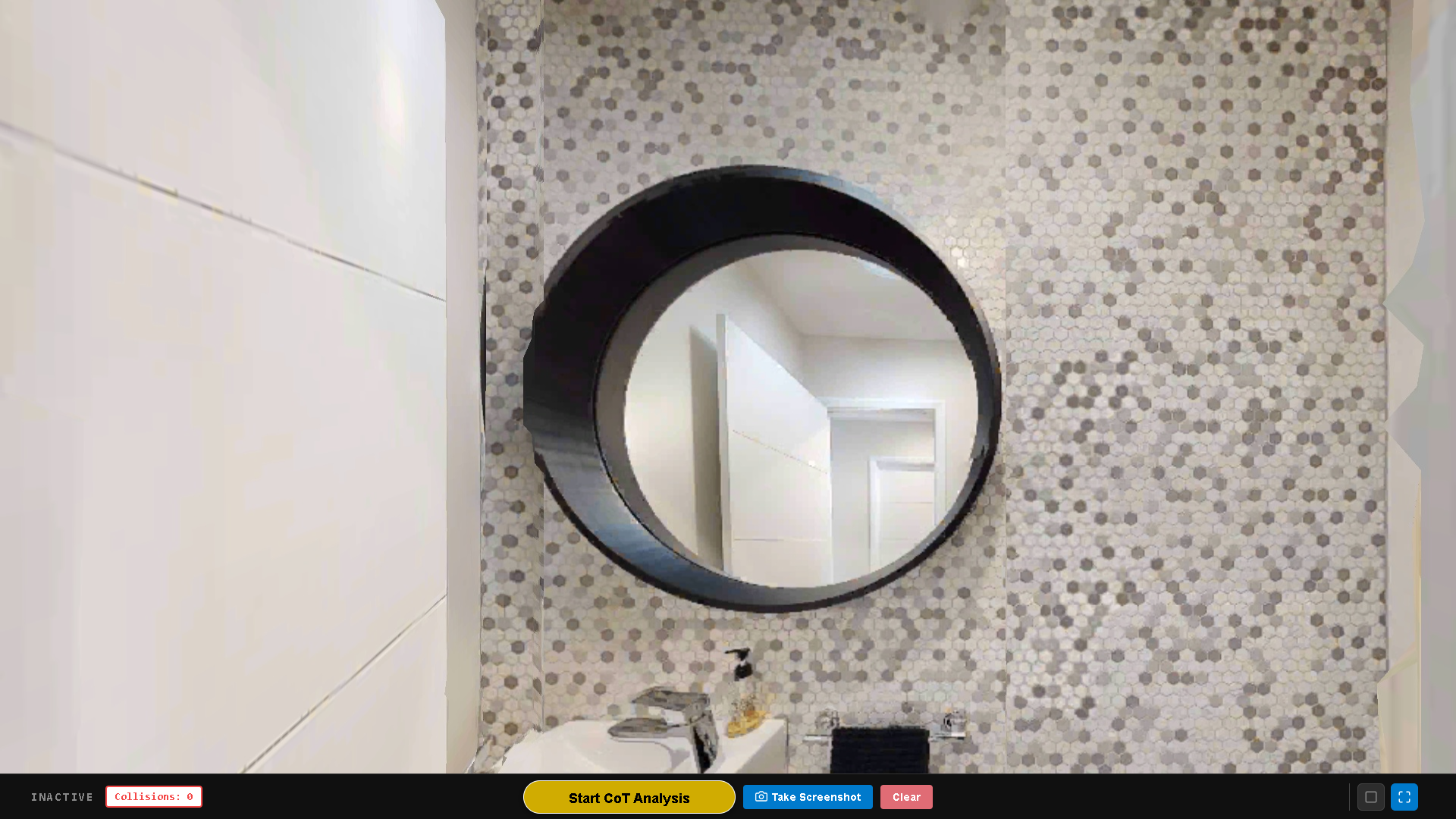}} 
  \subfloat[Exp 07]{\includegraphics[width=0.25\textwidth]{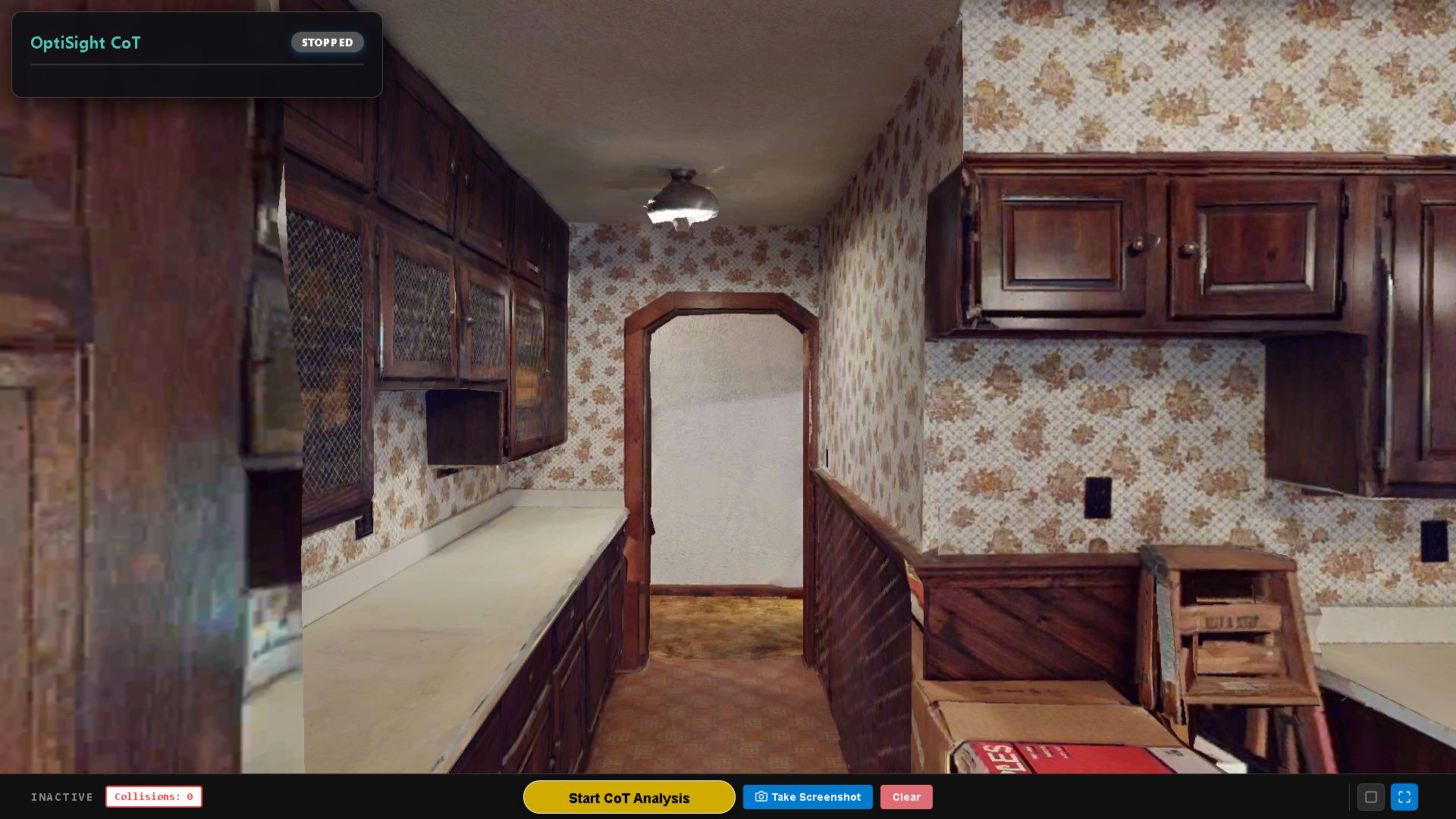}} 
  \subfloat[Exp 08]{\includegraphics[width=0.25\textwidth]{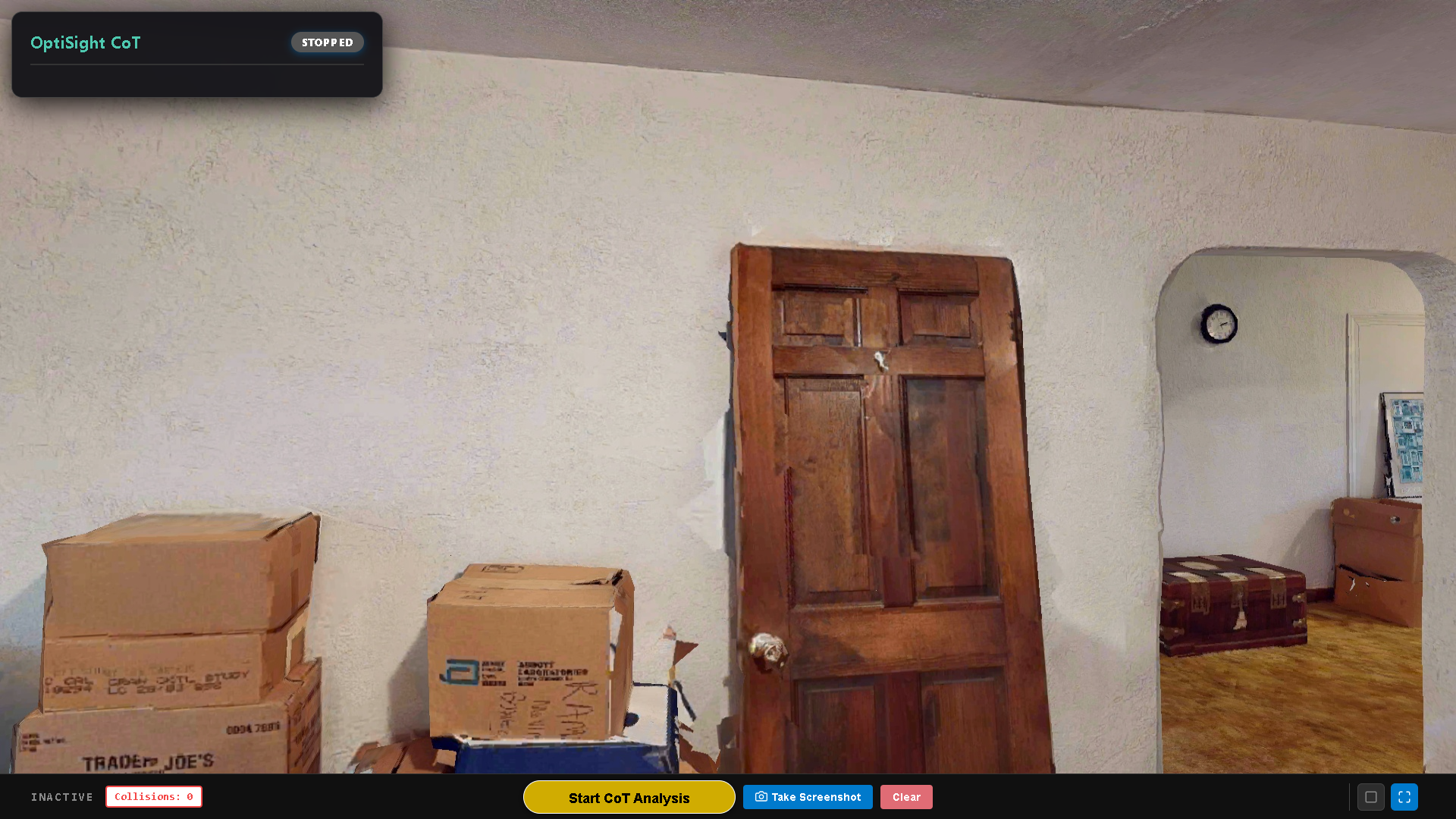}} 

  ~\\

  \subfloat[Exp 09]{\includegraphics[width=0.25\textwidth]{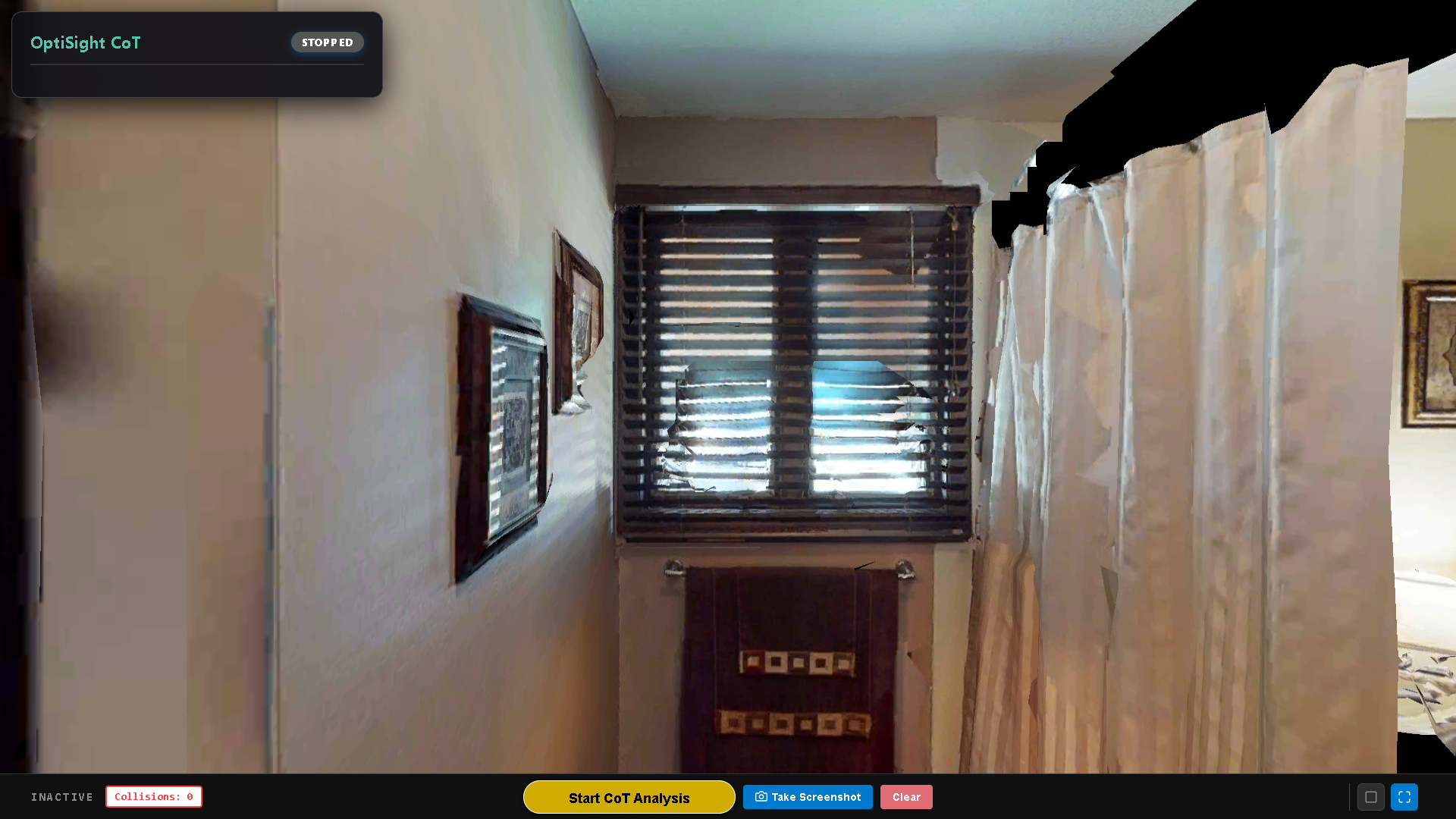}} 
  \subfloat[Exp 10]{\includegraphics[width=0.25\textwidth]{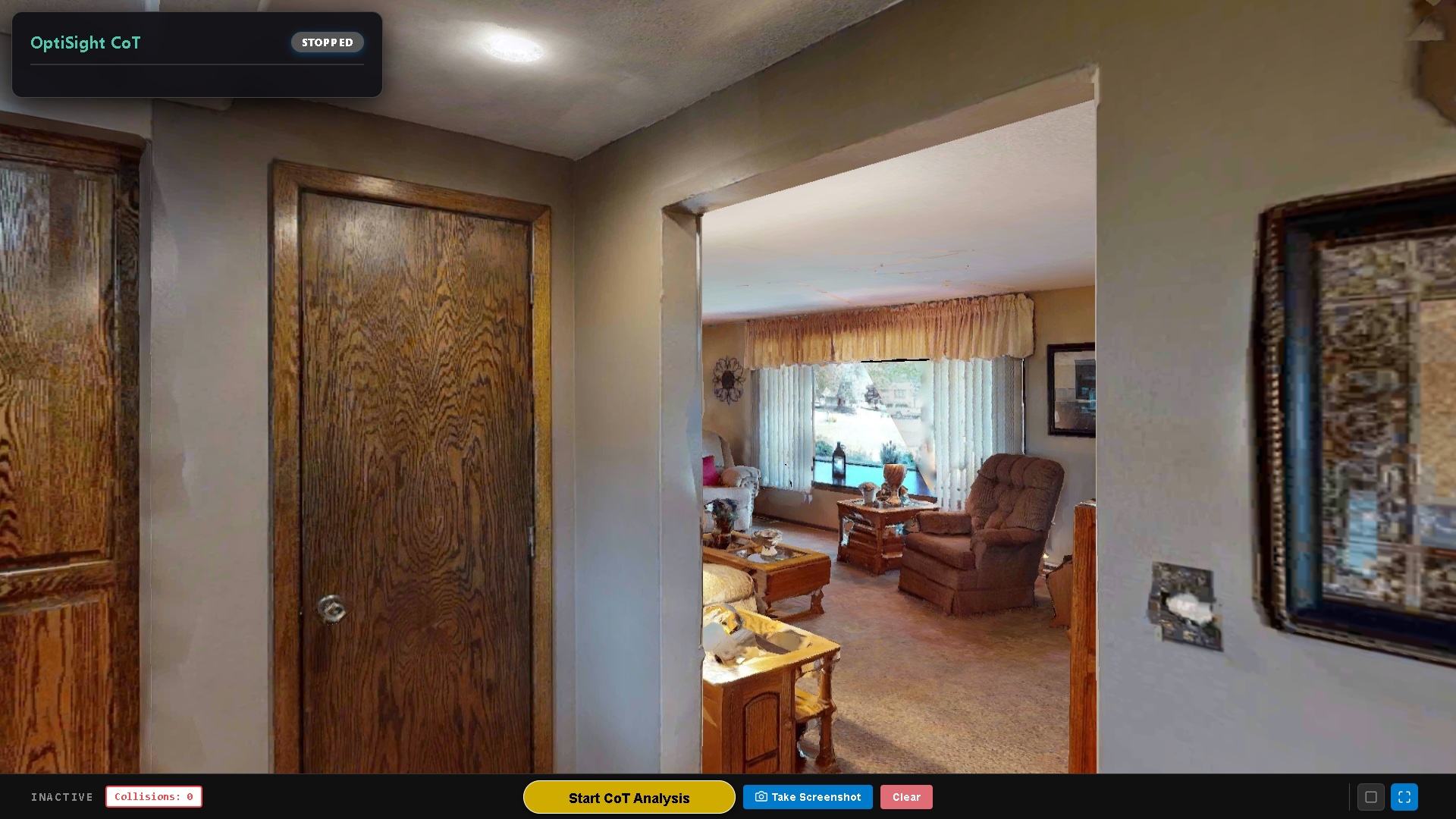}} 
  \subfloat[Exp 11]{\includegraphics[width=0.25\textwidth]{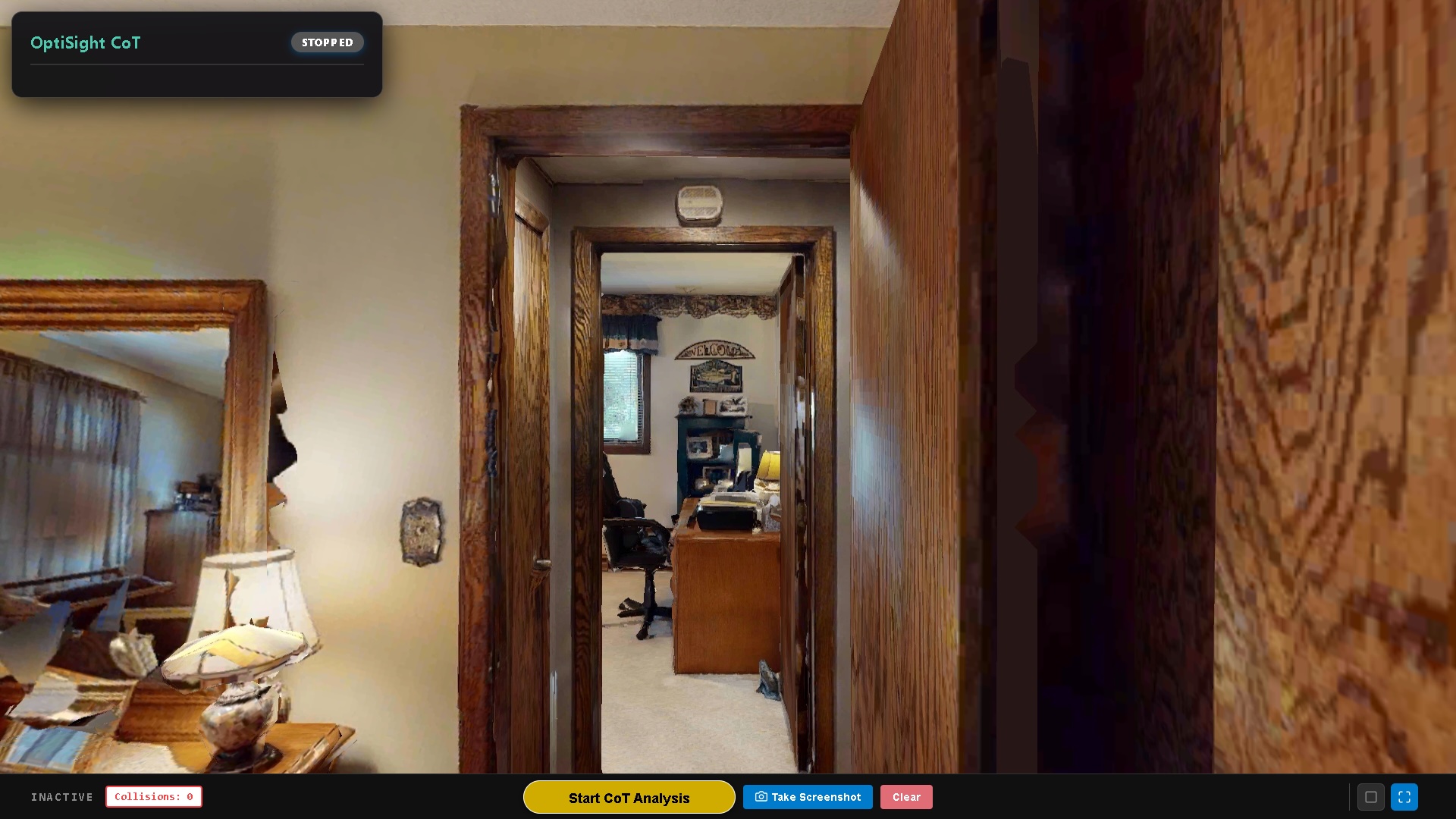}} 
  \subfloat[Exp 12]{\includegraphics[width=0.25\textwidth]{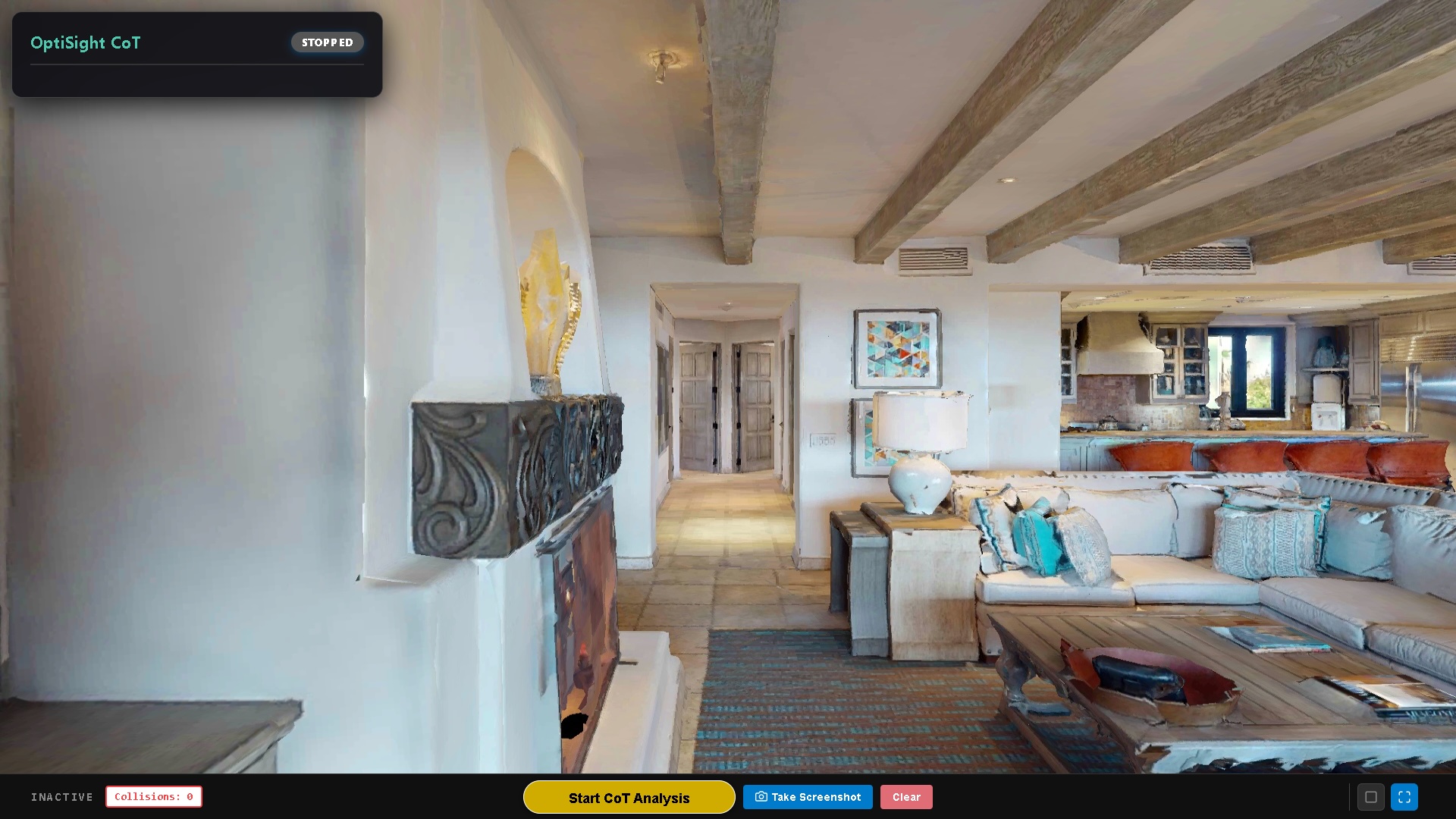}} 

  \caption{Initial frames across Experiments 01--12 indoor navigation scenarios used for the experimental evaluation of OptiSight. Each frame illustrates the initial observation provided to the agent before navigation begins.}
 
\label{fig:exps1}
\end{figure*}

Table~\ref{tab:final_comparison_1} summarizes the performance of OptiSight across the first 12 experiments. Overall, the framework successfully completes 8 out of 12 scenarios according to the defined mission-success criterion, with successful scenarios generally requiring only a small number of VLM requests and no collisions or recovery actions. Experiments~01--03, 07, 09, 11, and 12 achieve success rates between 60\% and 100\%, while Experiments~03, 07, and 09 achieve a 100\% success rate. In Figure~\ref{fig:exps1} we show the starting frame in the scenario.

The results also reveal a clear relationship between navigation difficulty and the amount of reasoning required. In the successful scenarios, OptiSight typically requires only one to three VLM requests, whereas the more challenging scenarios require substantially more interactions. For example, Experiments~05, 06, and 08 require 8, 6, and 6 VLM requests, respectively, and involve multiple recovery actions and collisions. These experiments also exhibit the longest execution times, ranging from 59.0 to 68.7\,s, indicating that the additional reasoning and recovery cycles are associated with increased navigation complexity.

\begin{table*}[ht]
\centering
\begin{tabular}{lcccccccccccc}
\toprule
\textbf{Metric} & \textbf{Exp 01} & \textbf{Exp 02} & \textbf{Exp 03} & \textbf{Exp 04} & \textbf{Exp 05} & \textbf{Exp 06} & \textbf{Exp 07} & \textbf{Exp 08} & \textbf{Exp 09} & \textbf{Exp 10} & \textbf{Exp 11} & \textbf{Exp 12}\\ 
\midrule
\textbf{Mission Success} & \cmark & \cmark & \cmark & \xmark & \xmark & \xmark & \cmark & \xmark & \cmark & \xmark & \cmark & \cmark \\ 
\textbf{Success Rate} & 80\% & 60\% & 100\% & 20\% & 0\% & 0\% & 100\% & 0\% & 100\% & 0\% & 80\% & 70\% \\ 
\textbf{Execution Time} & 31.5 s & 29.2 s & 24.7 s & 21.3 s & 59.0 s & 62.4 s & 31.6 s & 68.7 s & 32.1 s & 55.3 s & 19.0 s & 29.7 s \\ 
\textbf{Total Distance} & 5.03 m & 5.28 m & 4.00 m & 1.28 m & 8.07 m & 6.18 m & 4.5 m & 6.92 m & 2.5 m & 5.74 m & 2 m & 6.03 m\\ 
\textbf{Total Steps} & 64 & 62 & 50 & 35 & 111 & 118 & 50 & 95 & 44 & 71 & 32 & 57 \\ 
\textbf{Collisions} & 0 & 0 & 0 & 0 & 3 & 6 & 0 & 3 & 0 & 2 & 0 & 0 \\ 
\textbf{Recoveries} & 0 & 0 & 0 & 0 & 3 & 3 & 0 & 3 & 0 & 2 & 0 & 0 \\ 
\textbf{VLM Requests} & 2 & 2 & 2 & 3 & 8 & 6 & 2 & 6 & 3 & 2 & 1 & 1 \\ 
\textbf{Min Obstacle Dist} & 0.34 m & 0.27 m & 1.33 m & N/A & 0.95 m & 0.71 m & 1.67 m & 0.53 m & N/A & 0.76 m & 0.74 m & N/A \\ 
\bottomrule
\end{tabular}

\caption{Performance of OptiSight across Experiments 01--12 using Qwen3.5-2B for VLM reasoning. Results are averaged over 10 independent runs for each scenario.}
\label{tab:final_comparison_1}
\end{table*}

Obstacle avoidance is particularly reflected in the relationship between collisions and recovery actions. Experiments~05, 06, 08, and 10 are the only scenarios in which collisions are recorded, and each collision-prone experiment also triggers recovery actions. In contrast, all scenarios without collisions require no recovery procedure. This indicates that the recovery mechanism is effectively activated when the planned trajectory encounters an obstacle, allowing the system to reassess the environment rather than continuing along an invalid trajectory.

The variation in execution time and traveled distance further reflects the diversity of the evaluated scenarios. Successful experiments generally complete within 19.0--32.1\,s and traverse between 2.0 and 6.03\,m, although more complex layouts can require longer trajectories. In comparison, Experiments~05, 06, and 08 exhibit substantially longer execution times and trajectories, consistent with the additional obstacle-avoidance and recovery operations.

\begin{figure*}[ht]
  \centering
  
  \subfloat[Exp 13]{\includegraphics[width=0.25\textwidth]{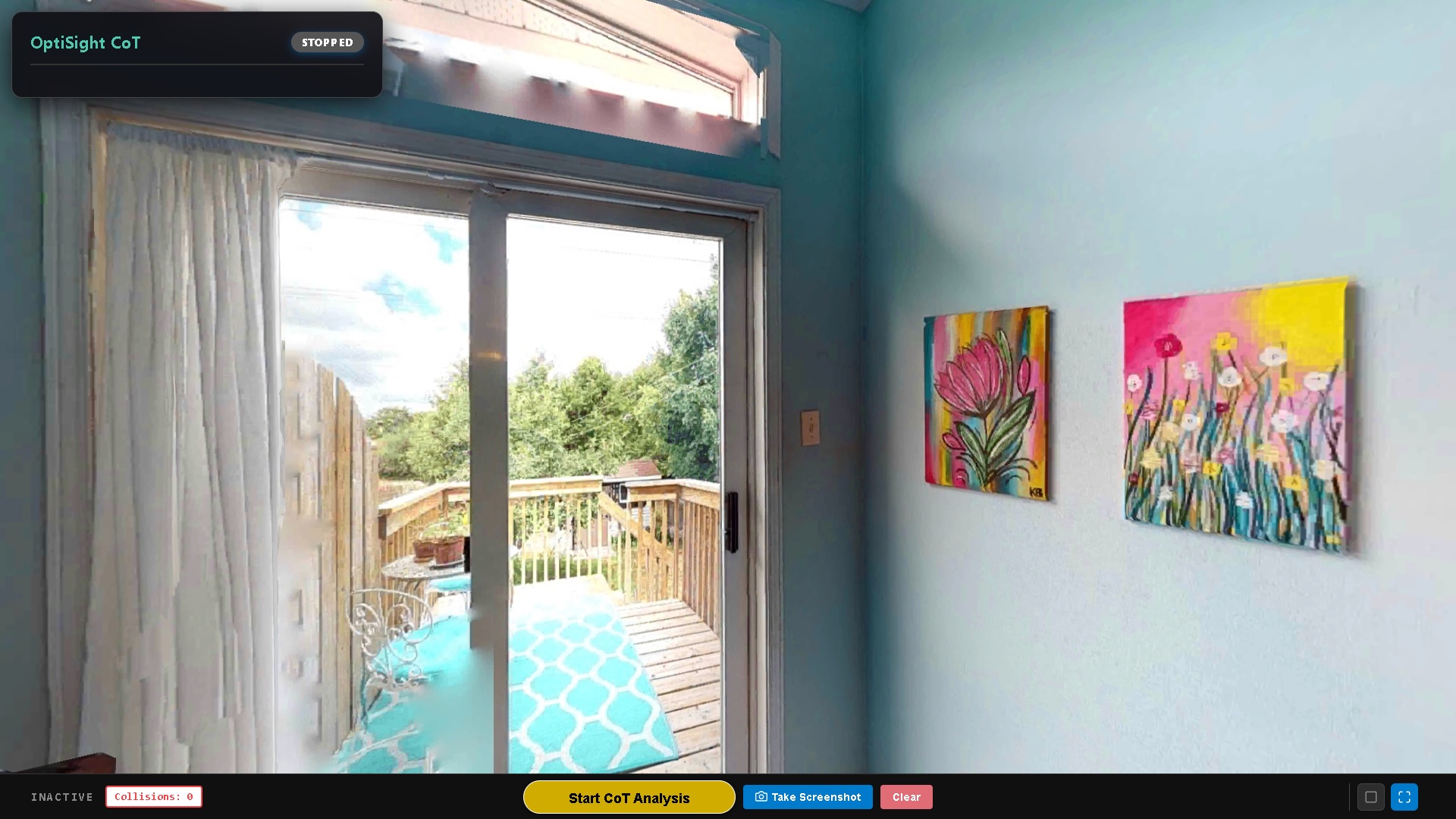}} 
  \subfloat[Exp 14]{\includegraphics[width=0.25\textwidth]{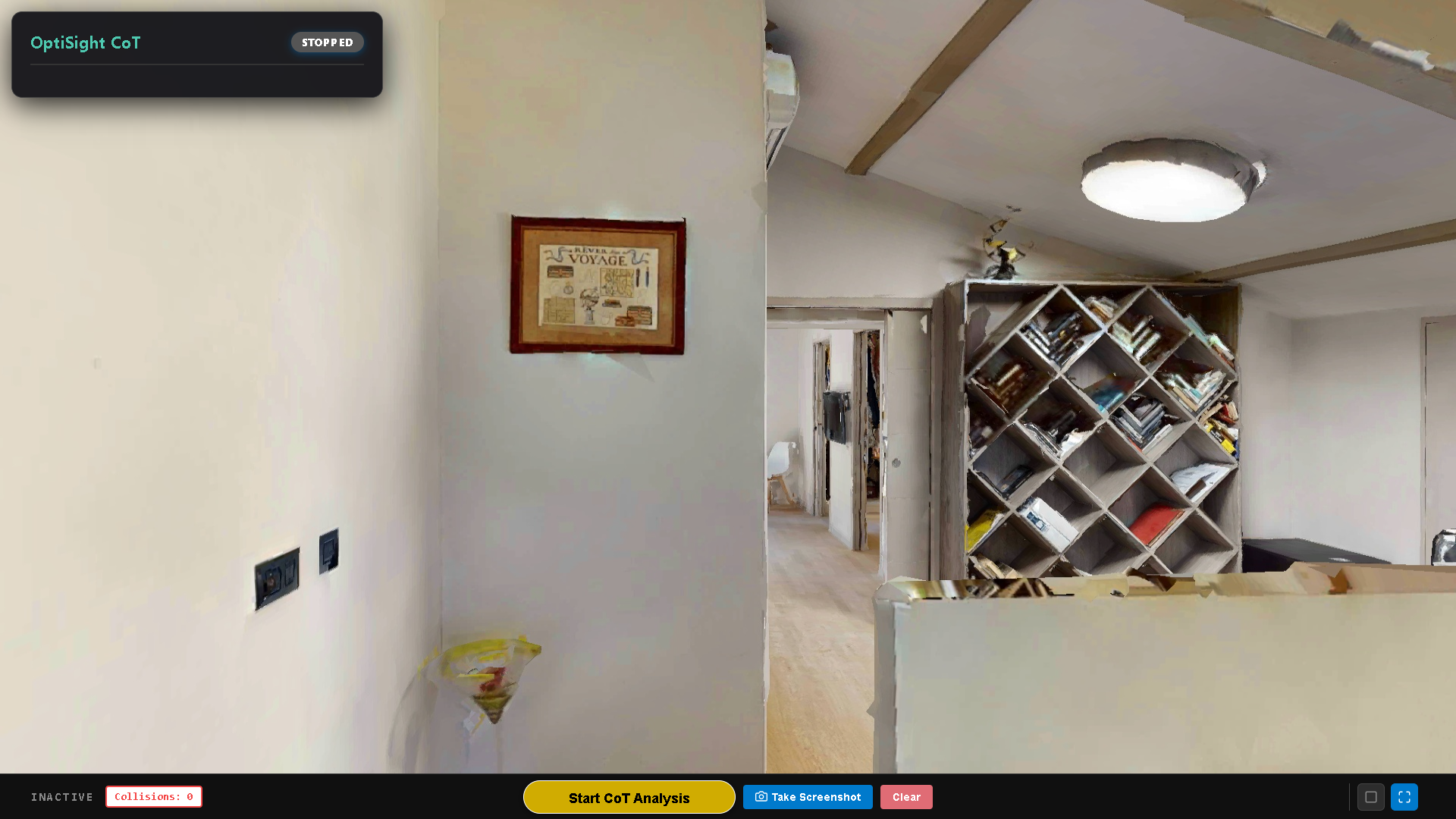}} 
  \subfloat[Exp 15]{\includegraphics[width=0.25\textwidth]{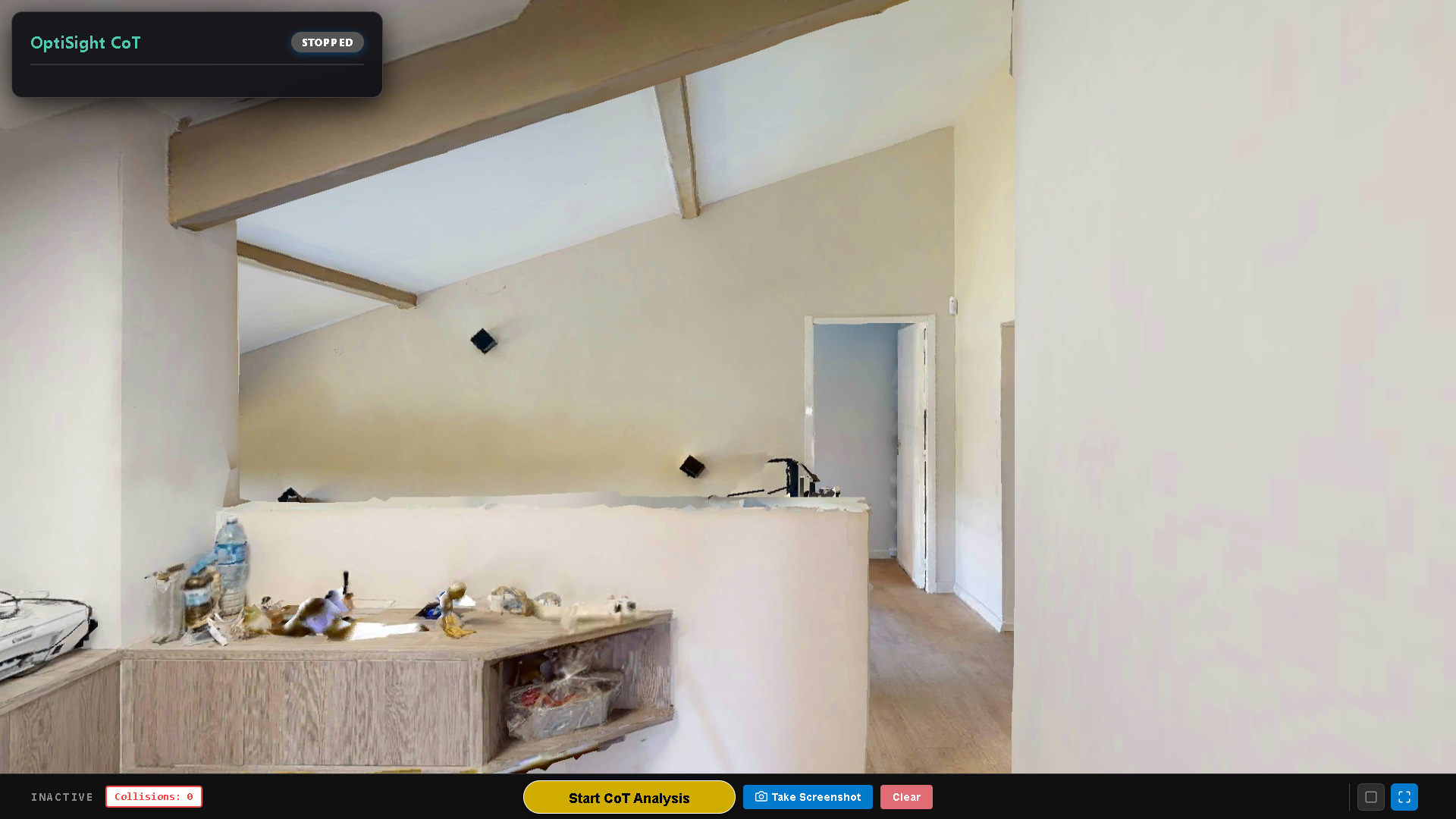}} 
  \subfloat[Exp 16]{\includegraphics[width=0.25\textwidth]{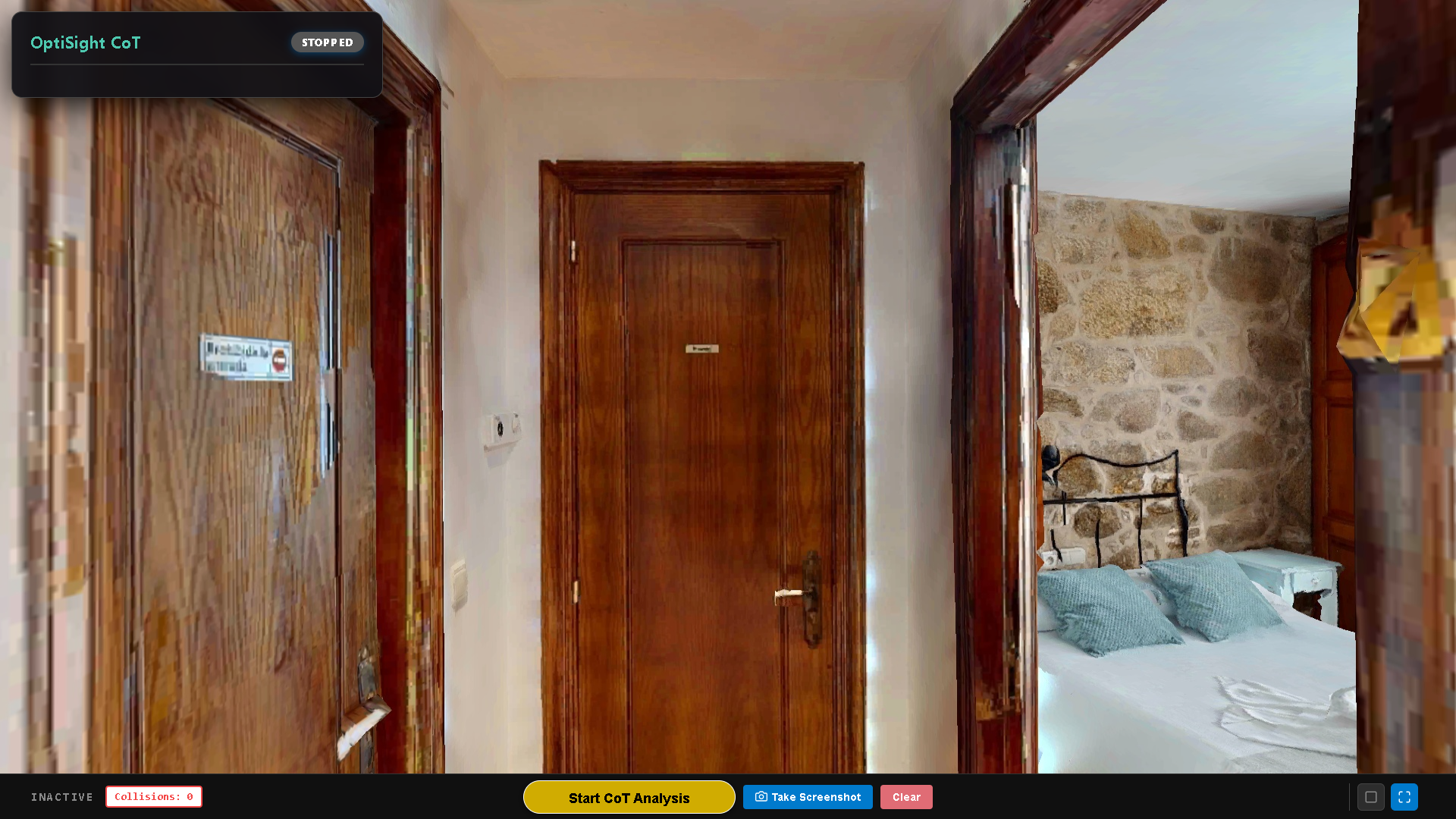}} 

  ~\\

  \subfloat[Exp 17]{\includegraphics[width=0.25\textwidth]{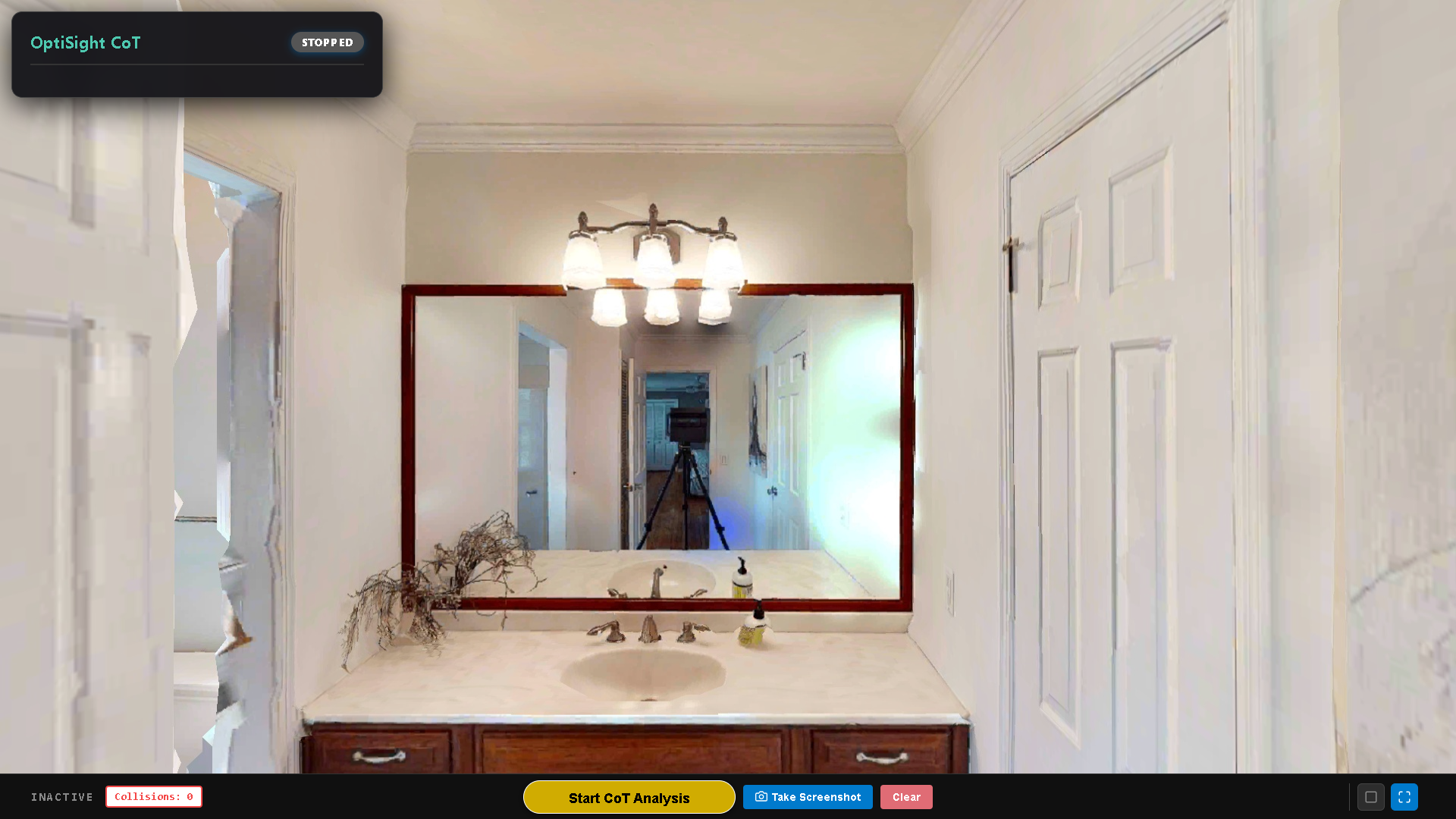}} 
  \subfloat[Exp 18]{\includegraphics[width=0.25\textwidth]{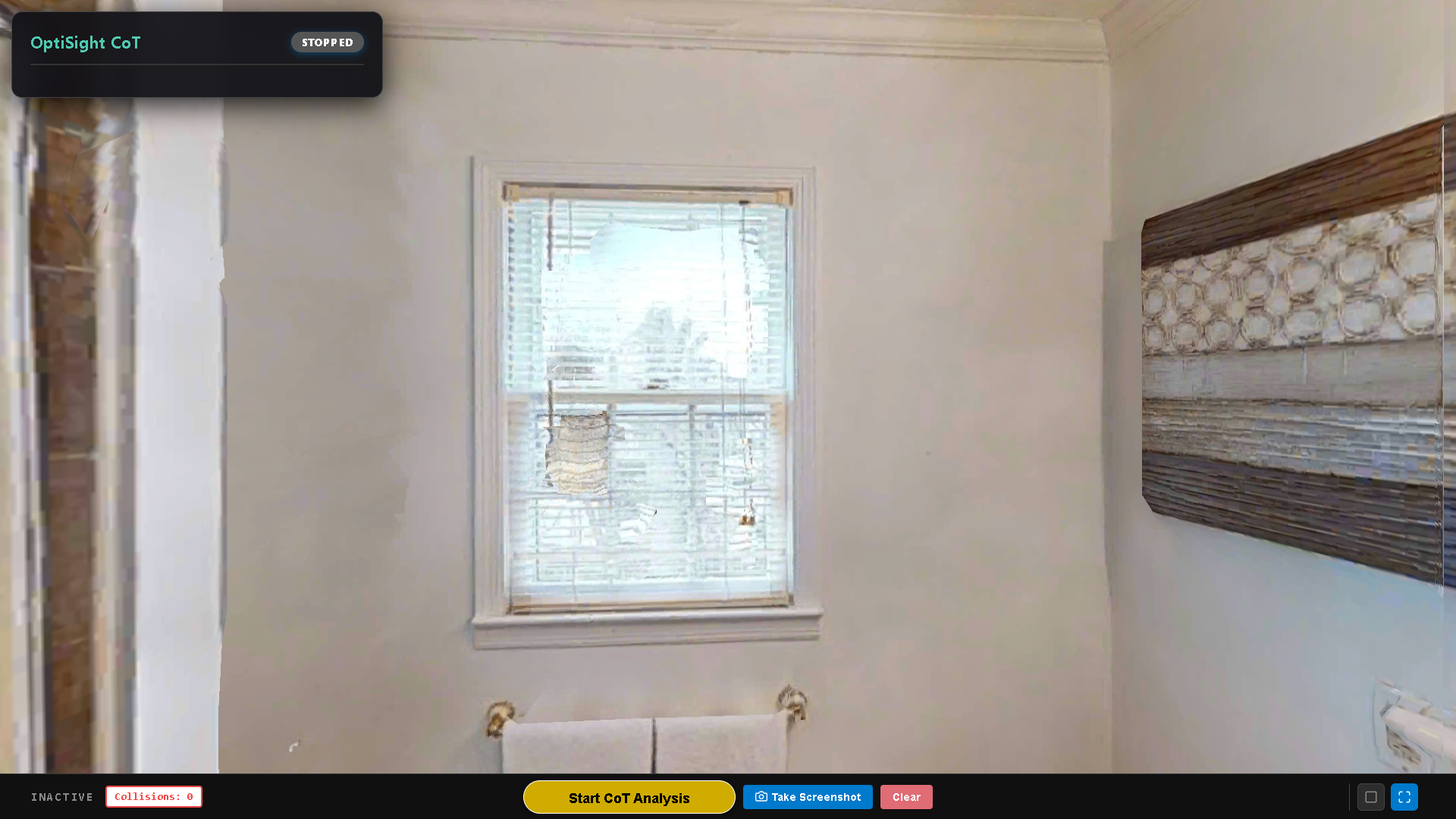}} 
  \subfloat[Exp 19]{\includegraphics[width=0.25\textwidth]{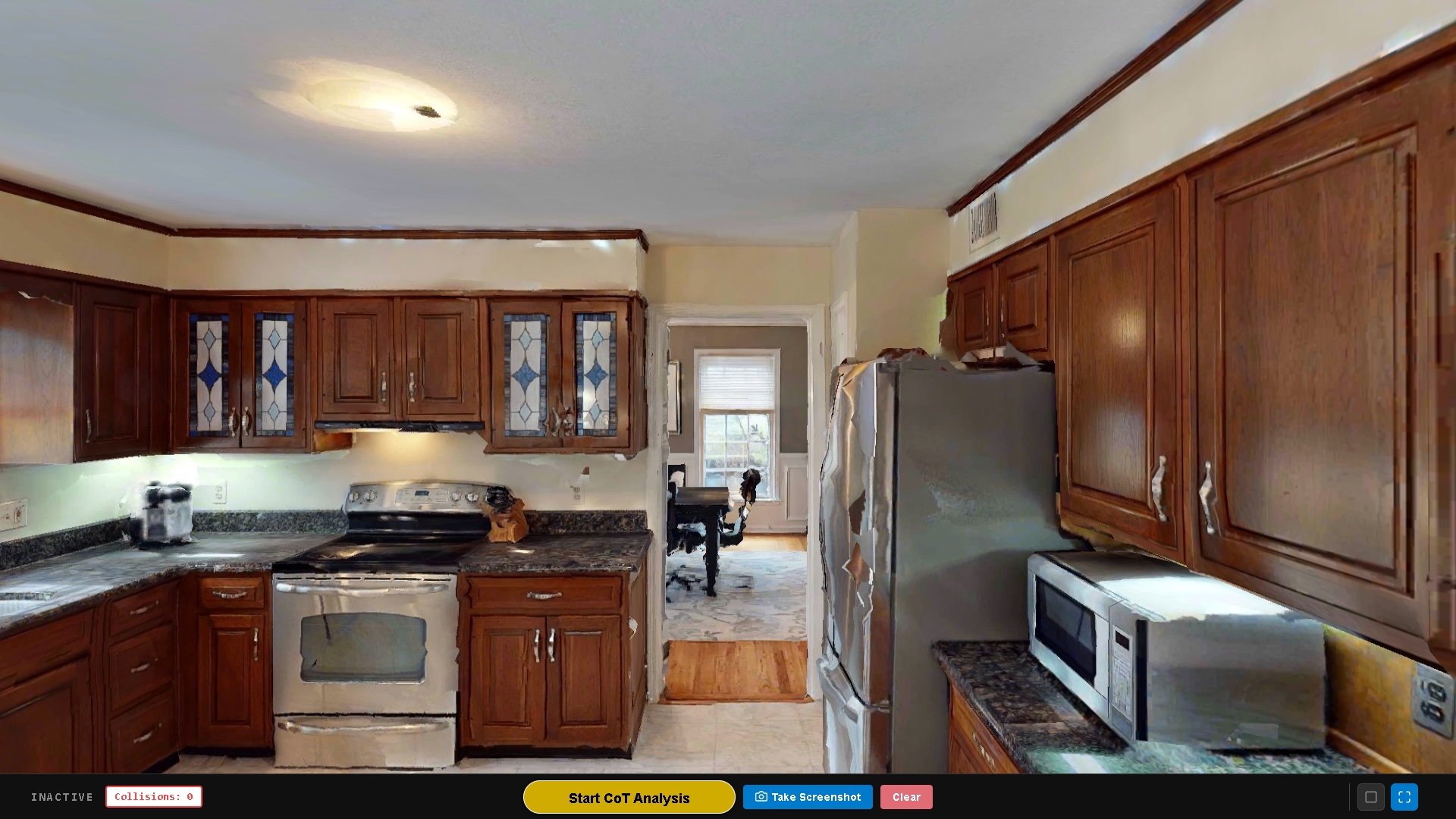}} 
  \subfloat[Exp 20]{\includegraphics[width=0.25\textwidth]{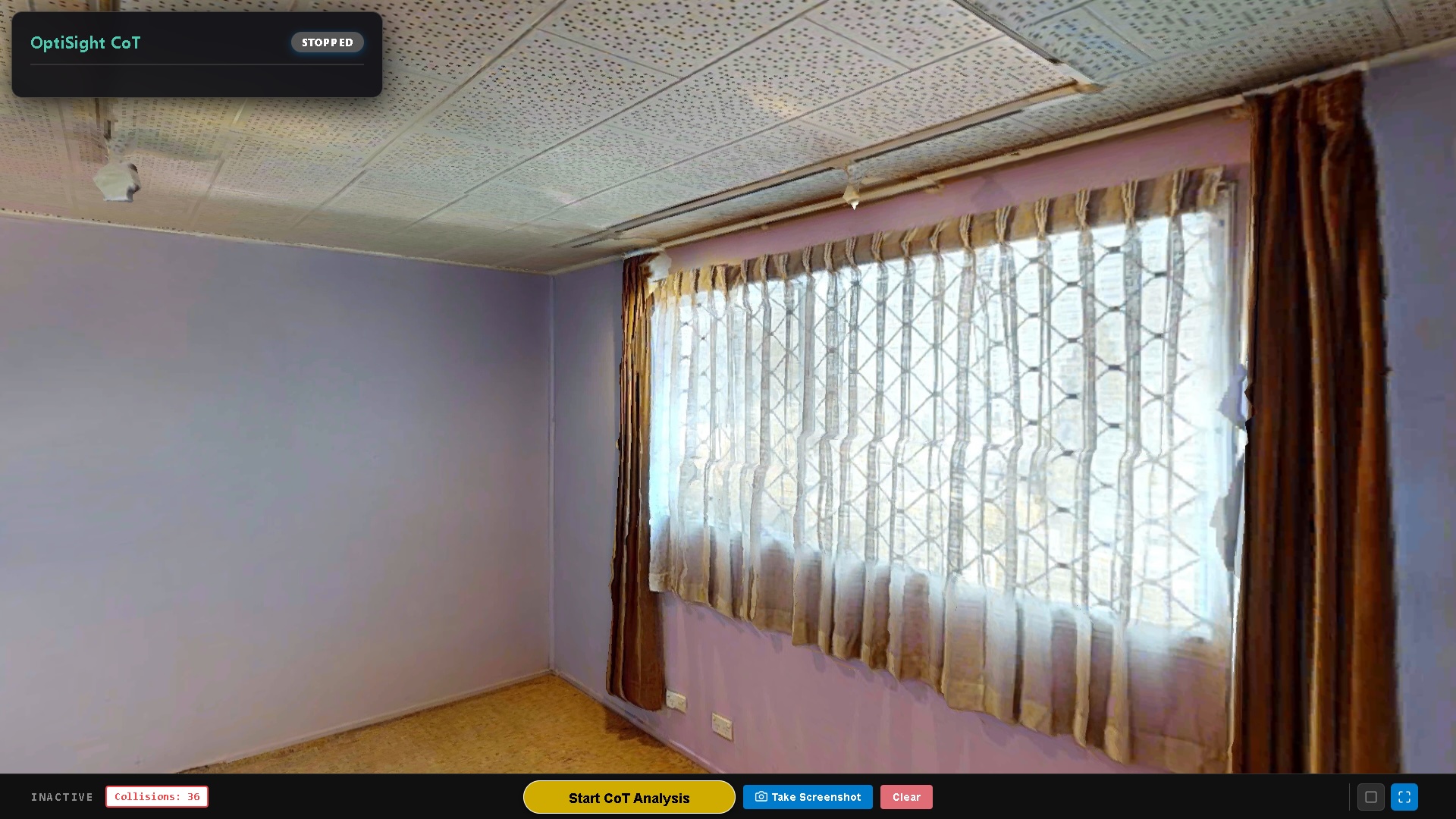}} 

  ~\\

  \subfloat[Exp 21]{\includegraphics[width=0.25\textwidth]{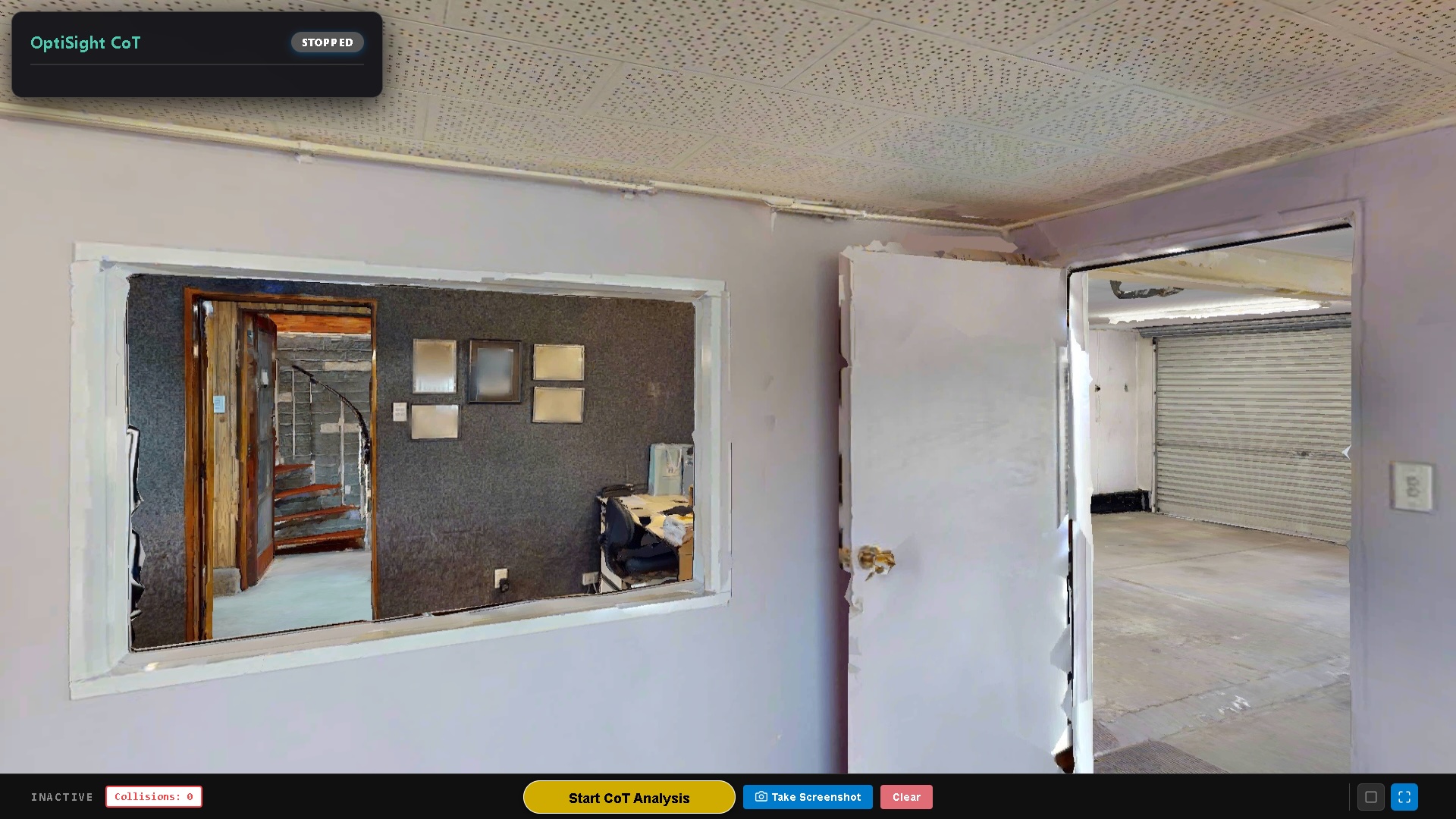}} 
  \subfloat[Exp 22]{\includegraphics[width=0.25\textwidth]{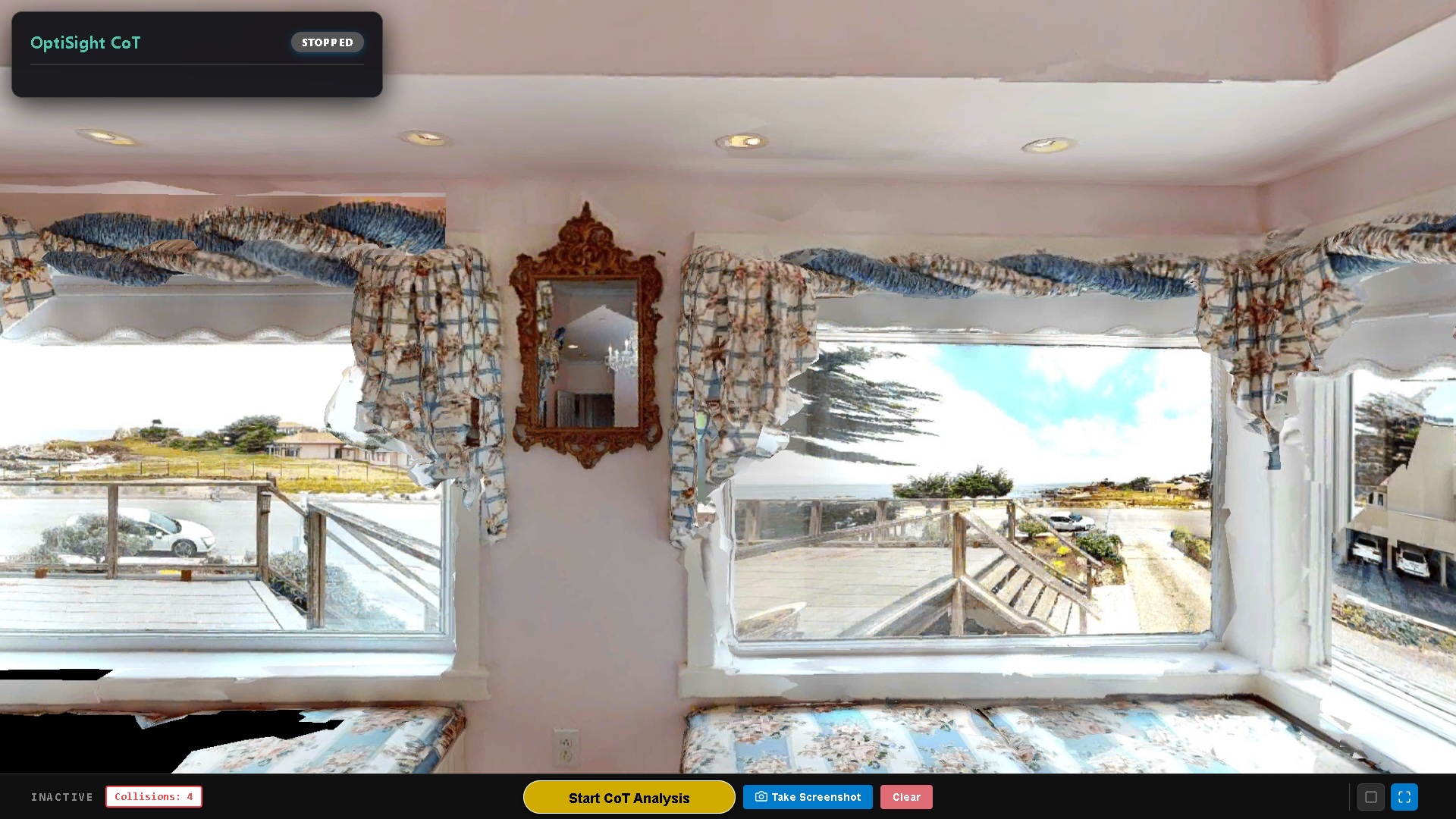}} 
  \subfloat[Exp 23]{\includegraphics[width=0.25\textwidth]{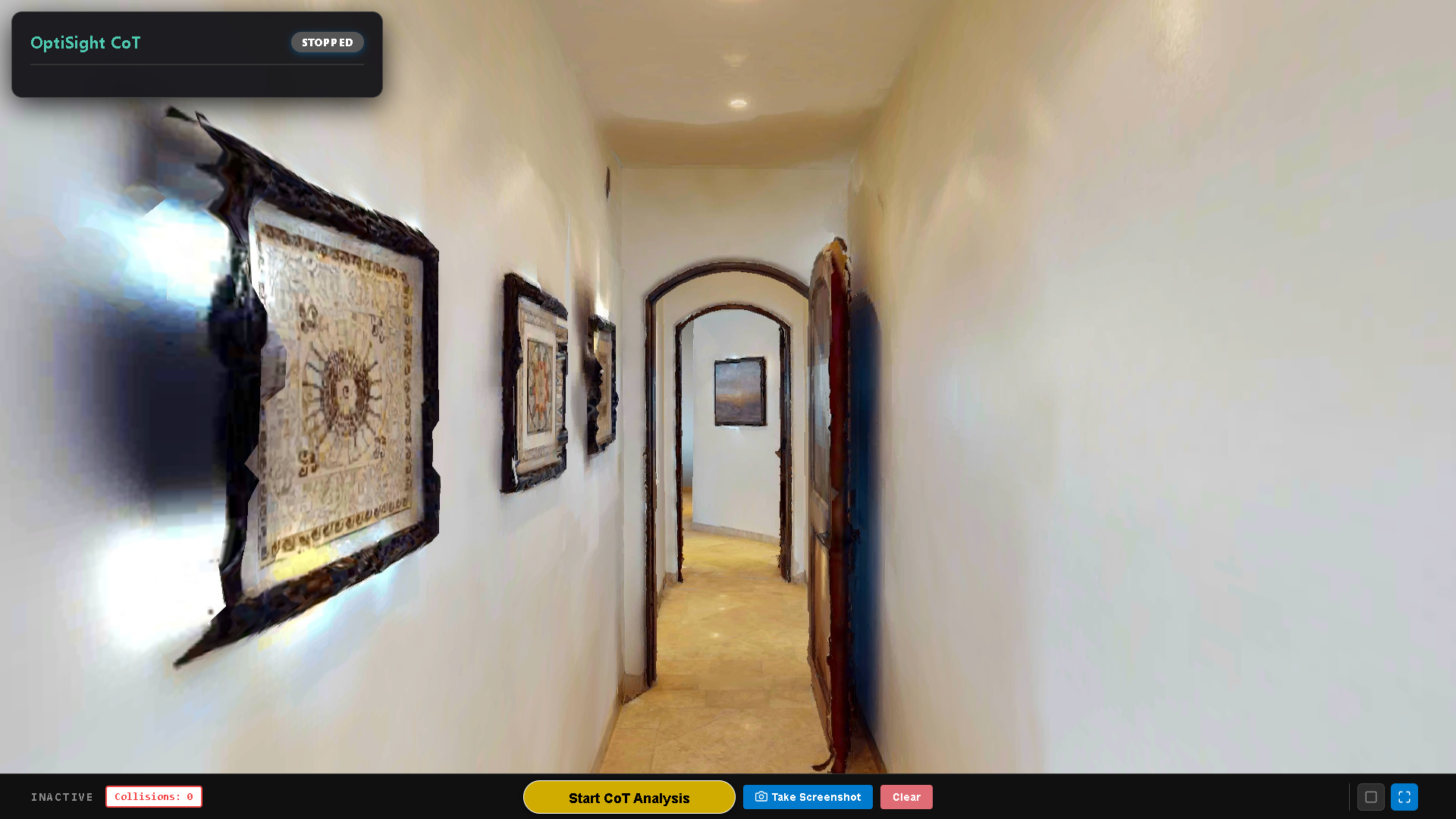}} 
  \subfloat[Exp 24]{\includegraphics[width=0.25\textwidth]{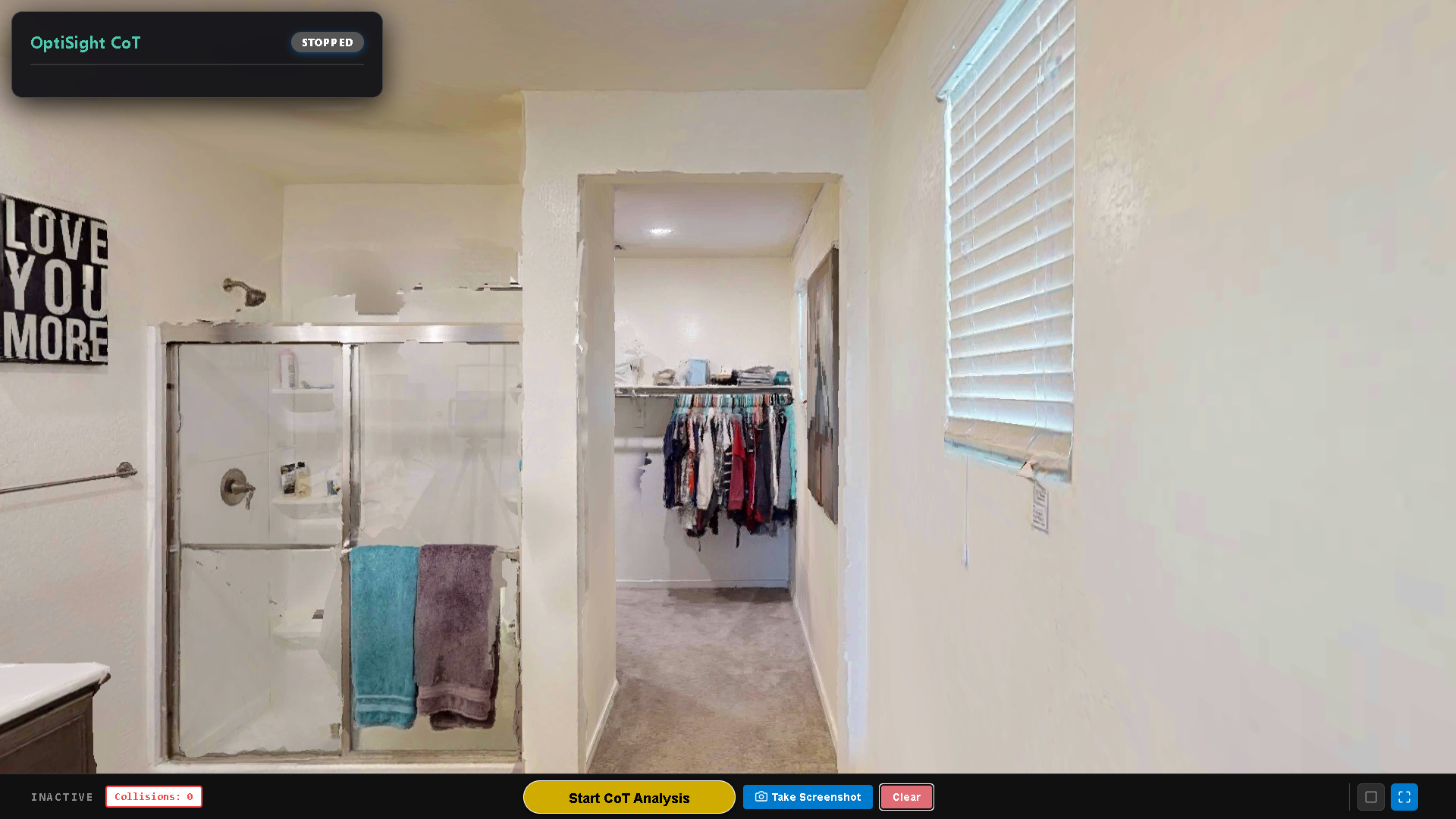}} 

  \caption{Initial frames across Experiments 13--24 indoor navigation scenarios used for the experimental evaluation of OptiSight. Each frame illustrates the initial observation provided to the agent before navigation begins.}
 
\label{fig:exps2}
\end{figure*}

Table~\ref{tab:final_comparison_2} reports the performance of OptiSight across Experiments~13--24. Overall, 7 of the 12 scenarios satisfy the defined mission-success criterion, with four experiments (19, 20, 23, and 24) achieving a 100\% success rate. Experiments~15 and 18 also demonstrate relatively high success rates of 80\% and 70\%, respectively, indicating that the framework can successfully complete several challenging scenarios despite the increased execution time observed for this group of experiments. In Figure~\ref{fig:exps2} we show the starting frame in the scenario.

\begin{table*}[ht]
\centering
\begin{tabular}{lcccccccccccc}
\toprule
\textbf{Metric} & \textbf{Exp 13} & \textbf{Exp 14} & \textbf{Exp 15} & \textbf{Exp 16} & \textbf{Exp 17} & \textbf{Exp 18} & \textbf{Exp 19} & \textbf{Exp 20} & \textbf{Exp 21} & \textbf{Exp 22} & \textbf{Exp 23} & \textbf{Exp 24}\\ 
\midrule
\textbf{Mission Success} & \xmark & \cmark & \cmark & \xmark & \xmark & \cmark & \cmark & \xmark & \xmark & \xmark & \cmark & \cmark \\ 
\textbf{Success Rate} & 0\% & 60\% & 80\% & 0\% & 0\% & 70\% & 100\% & 100\% & 30\% & 0\% & 100\% & 100\% \\ 
\textbf{Execution Time} & 97.3 s & 85.4 s & 61.0 s & 87.6 s & 98.9 s & 91.8 s & 73.1 s & 84.0 s & 94.2 s & 118.1 s & 46.5 s & 67.7 s \\ 
\textbf{Total Distance} & 5.59 m & 7.96 m & 4.75 m & 5.37 m & 5.12 m & 1.49 m & 5.0 m & 4.5 m & 4.5 m & 4.48 m & 3.75 m & 3.49 m \\ 
\textbf{Total Steps} & 72 & 88 & 67 & 72 & 77 & 68 & 55 & 70 & 79 & 76 & 40 & 61 \\ 
\textbf{Collisions} & 2 & 1 & 0 & 2 & 2 & 0 & 0 & 0 & 0 & 4 & 0 & 0 \\ 
\textbf{Recoveries} & 2 & 1 & 0 & 2 & 2 & 0 & 0 & 0 & 0 & 2 & 0 & 0 \\ 
\textbf{VLM Requests} & 2 & 1 & 1 & 2 & 3 & 3 & 1 & 4 & 1 & 3 & 1 & 1 \\ 
\textbf{Min Obstacle Dist} & 0.7 m & 0.5 m & 0.34 m & 1.37 m & 0.68 m & 0.59 m & 0.42 m & 0.56 m & 0.56 m & 0.15 m & 0.52 m & 0.25 \\ 
\bottomrule
\end{tabular}%

\caption{Performance of OptiSight across Experiments 13--24 using Moondream2-2B for VLM reasoning. Results are averaged over 10 independent runs for each scenario.}
\label{tab:final_comparison_2}
\end{table*}

Compared with the first set of experiments, these scenarios generally require longer execution times, with values ranging from 46.5\,s to 118.1\,s. The longest execution times are observed in Experiments~13, 17, and 22, which require 97.3, 98.9, and 118.1\,s, respectively. These longer durations are not directly associated with a larger number of VLM requests, since most experiments require only one to three requests. This behavior highlights the benefit of the state-driven architecture, where extended execution can result primarily from geometric navigation, obstacle avoidance, or repeated scanning rather than continuous VLM inference.

Obstacle interactions provide a clearer explanation for several unsuccessful scenarios. Collisions are reported in Experiments~13, 14, 16, 17, and 22, and each of these experiments also triggers recovery actions. In particular, Experiment~22 records four collisions and two recovery procedures and has the lowest minimum obstacle clearance in the table, at only 0.15,m. Experiments~13, 16, and 17 similarly combine collisions with unsuccessful missions, suggesting that navigation becomes more challenging when obstacles are encountered at close proximity. In contrast, Experiments~19--21 and 23--24 complete without collisions or recovery actions, despite requiring between one and four VLM requests.

The minimum obstacle distance further illustrates the range of geometric conditions encountered in these scenarios. Successful experiments can operate with relatively small clearances, such as 0.25\,m in Experiment~24 and 0.34\,m in Experiment~15, while Experiment~22 reaches a minimum clearance of only 0.15\,m and experiences multiple collisions. This suggests that maintaining sufficient clearance remains an important factor for reliable navigation, particularly in scenarios involving constrained passages or closely positioned obstacles.

An important observation is that the number of VLM requests remains relatively low across all experiments, ranging from one to four calls. Several successful scenarios, including Experiments~19, 23, and 24, require only a single VLM request. This result is consistent with the state-driven design of OptiSight, in which semantic reasoning is invoked only at key decision points while geometric planning and waypoint navigation are handled independently.

\section{Conclusions}
We presented \textbf{OptiSight}, a hybrid autonomous navigation framework that combines selective Vision-Language Model reasoning with deterministic geometric navigation. By integrating a finite-state machine, Grounded-SAM-based target grounding, and visual-to-3D geometric projection, OptiSight translates high-level semantic instructions into executable navigation actions while avoiding continuous VLM inference and computationally expensive mapping pipelines.

Evaluation in AI Habitat across 24 indoor navigation scenarios demonstrates that OptiSight can successfully handle diverse navigation conditions, including obstacle avoidance, semantic ambiguity, partial observability, and challenging viewpoints, while requiring only a limited number of VLM queries. These results highlight the potential of combining semantic reasoning with lightweight geometric control for efficient embodied navigation.

Future work will focus on extending the framework to a broader range of navigation tasks and validating OptiSight on physical robotic platforms.

\bibliographystyle{IEEEtran}
\bibliography{references}

\end{document}